\documentclass[runningheads]{llncs}

\usepackage{eccv}

\usepackage{booktabs, multirow}
\usepackage{eccvabbrv}
\usepackage{amsfonts}       
\usepackage{nicefrac}       
\usepackage{rotating}
\usepackage{microtype}      
\usepackage{xcolor}         
\usepackage{amsmath}
\usepackage{amssymb}
\usepackage{url}
\usepackage{microtype}
\usepackage{graphicx}
\usepackage{booktabs} %
\usepackage{colortbl}
\usepackage{tabulary}
\usepackage{adjustbox}
\usepackage{multirow}
\usepackage{makecell} 
\usepackage{caption}
\usepackage{subcaption}
\usepackage{enumitem}
\usepackage{xcolor}
\usepackage{wrapfig}
\usepackage{stackengine}
\usepackage{comment}
\usepackage{mathtools}
\usepackage{url}
\usepackage{microtype}
\usepackage{graphicx}
\usepackage{booktabs} %
\usepackage{colortbl}
\usepackage{tabulary}
\usepackage{adjustbox}
\usepackage{multirow}
\usepackage{makecell} 
\usepackage{caption}
\usepackage{subcaption}
\usepackage{enumitem}
\usepackage{xcolor}
\usepackage{multirow}
\usepackage{caption}
\usepackage{tabularx}
\usepackage{subcaption}
\usepackage{wrapfig}
\usepackage{stackengine}
\usepackage{comment}
\usepackage{mathtools}
\usepackage{xcolor} 
\usepackage{pifont} 
\usepackage{tikz}
\usepackage{graphicx} 
\usepackage{booktabs} 
\usepackage{tabularx} 
\usepackage{ragged2e} 
\usepackage{comment}
\usepackage{lipsum}
\usepackage{minitoc}
\usepackage{amsmath,amssymb} 
\usepackage{color}
\usepackage{graphicx}
\usepackage{booktabs}
\usepackage{hyperref}
\usepackage[accsupp]{axessibility}  
\usepackage{tcolorbox}
\tcbuselibrary{breakable}
\usepackage{longtable}
\usepackage{tabularray}
\usepackage{framed}
\usepackage{subcaption}
\usepackage{caption}
\usepackage{marvosym}
\tcbuselibrary{listings, skins} 

\newcommand\hb{ \rowcolor{teal!7}}

\newcommand{\red}[1]{\textcolor{red}{#1}}

\newcommand\our{{VisionPrefer}}
\newcommand\ourscore{{VP-Score}}
\newcommand\ourscorep{{VP-Score-P}}
\newcommand\ourscorea{{VP-Score-A}}
\newcommand\ourscoref{{VP-Score-F}}
\newcommand\ourscoreh{{VP-Score-H}}

\newcommand\gptv{\texttt{GPT-4 Vision}}
\newcommand\gpt{\texttt{GPT-4}}
\newcommand\gemini{\texttt{Gemini pro Vision}}
\newcommand\llava{\texttt{LLaVA 1.6-34B}}
\newcommand\ttname{Multimodal Large Language Model is a Human-Aligned Annotator for Text-to-Image Generation}
\newcommand{\cmark}{{\color{blue}\ding{51}}}%
\newcommand{\xmark}{{\color{red}\ding{55}}}%
\newcommand\hc{ \rowcolor{teal!20}}

\usepackage{hyperref}
\usepackage{orcidlink}
\usepackage[utf8]{inputenc} 

\usepackage{hyperref}
\begin{document}
\title{\ttname{}} 

\titlerunning{Abbreviated paper title}

\author{Xun Wu\inst{1,2}\thanks{~Contribution during internship at Microsoft. \textsuperscript{\Letter} Corresponding Authors.} \and
Shaohan Huang\inst{1}\textsuperscript{\Letter} \and Furu Wei\inst{1}}
\authorrunning{W.~Author et al.}
\institute{Microsoft Research Asia, Beijing, China \and
Tsinghua University, Beijing, China \\
\email{\{v-wuxun, shaohanh, fuwei\}@microsoft.com}\\
}

\maketitle

\begin{figure}
\centering
\includegraphics[width=\linewidth]{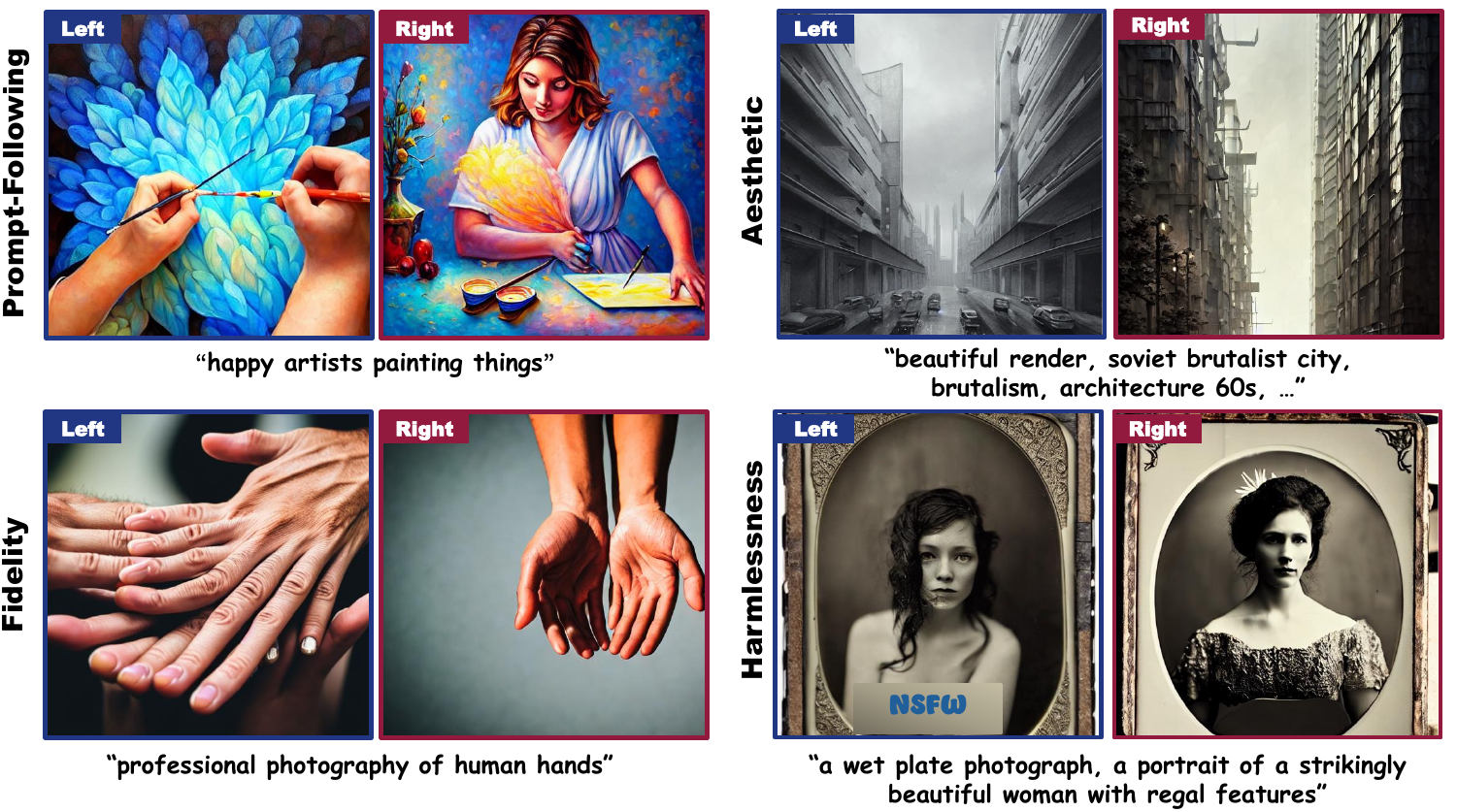}
\vspace{-6mm}
\caption{Fine-grained feedback from multimodal large language model help to yield more human-preferred images. Left: Output generated by the baseline text-to-image generative model. Right: Output generated by the baseline model optimized on our preference dataset \our{}. We illustrate improvements in generation quality across four aspects: \textbf{Prompt-Following}, \textbf{Aesthetic}, \textbf{Fidelity} and \textbf{Harmlessness}. See Appendix for more examples.}
\label{fig:top}
\vspace{-7mm}
\end{figure}

\begin{abstract}
Recent studies have demonstrated the exceptional potentials of leveraging human preference datasets to refine text-to-image generative models, enhancing the alignment between generated images and textual prompts. 
Despite these advances, current human preference datasets are either prohibitively expensive to construct or suffer from a lack of diversity in preference dimensions, resulting in limited applicability for instruction tuning in open-source text-to-image generative models and hinder further exploration.
To address these challenges and promote the alignment of generative models through instruction tuning, we leverage multimodal large language models to create \textbf{\our{}}, a high-quality and fine-grained preference dataset that captures multiple preference aspects. We aggregate feedback from AI annotators across four aspects: prompt-following, aesthetic, fidelity, and harmlessness to construct \our{}.
To validate the effectiveness of \our{}, we train a reward model \ourscore{} over \our{} to guide the training of text-to-image generative models and the preference prediction accuracy of \ourscore{} is comparable to human annotators. 
Furthermore, we use two reinforcement learning methods to supervised fine-tune generative models to evaluate the performance of \our{}, and extensive experimental results demonstrate that \our{} significantly improves text-image alignment in compositional image generation across diverse aspects, e.g., aesthetic, and generalizes better than previous human-preference metrics across various image distributions. Moreover, \our{} indicates that the integration of AI-generated synthetic data as a supervisory signal is a promising avenue for achieving improved alignment with human preferences in vision generative models.


\end{abstract}

\section{Introduction}


Text-to-image generative models~\cite{saharia2022photorealistic,rombach2022high,ramesh2022hierarchical,song2020denoising} have experienced rapid advancements in recent years. For example, large-scale text-to-image diffusion models, exemplified by Imagen~\cite{imagen} and DALL$\cdot$E2~\cite{ramesh2022hierarchical}, have demonstrated the capability to generate high-quality and creative images when provided with novel textual prompts.
Unfortunately, despite recent progress, current generative models still face a challenge: they may generate incorrect or unsafe content that deviates from human preferences, such as awkward limb and facial expression combinations. Additionally, due to noise in pre-training datasets, this often results in a misalignment between the semantics of generated images and their corresponding textual prompts~\cite{prabhudesai2023aligning}.

Reinforcement Learning from Human Feedback (RLHF)~\cite{ouyang2022training,stiennon2020learning,bai2022training} has proven effective in aligning Large Language Models (LLMs) with human preferences. Inspired by this success, RLHF has been applied to the supervised fine-tuning of text-to-image generative models, particularly diffusion models, by utilizing high-quality human preference data~\cite{lee2023aligning,black2023training,clark2023directly,prabhudesai2023aligning} and shows promising alignment results.
%


Preference data plays a pivotal role in the development of aligning generative models with text prompts. Unfortunately, in the area of text-to-image diffusion, collecting data from samples that demonstrate desired characteristics like aesthetics and fairness is not only a daunting and costly task but is also vulnerable to inherent biases. Another source of bias is in text prompts. User-generated prompts frequently adopt a structured format, comprising a descriptive passage supplemented by stylistic adjectives.However, these stylistic terms often contain contradictions, making it harder for human annotators to understand.
Besides, existing preference benchmarks, such as HPS v2~\cite{hps,hpsv2} and Pick-a-Pic~\cite{kirstain2023pick}, are either limited on alignment aspects only or in short of meticulous preference annotations. Drawing inspiration from recent research utilizing AI-generated data as training supervise signal, we pose the following question:

\begin{tcolorbox}[colback=gray!20, colframe=gray!50, sharp corners, center title]
\centering
\textit{\small Can Multimodal Large Language Models act as a Human-Aligned Annotator for Text-to-Image Generation?}
\end{tcolorbox}

Given the question mentioned above, we have contemplated whether multimodal large language models (MLLMs)~\cite{liu2023improved} can serve as human-aligned annotators. These MLLMs, trained on vast amounts of text and text-image pairs, have already demonstrated formidable capabilities on image understanding. To this end, we introduce \our{}, a publicly available AI-generated dataset that features millions of finely-grained human preferences concerning model-generated images.
Compared with existing human preference datasets, \our{} offers the following benefits:
(i) \textbf{Scalability}: As shown in Table~\ref{Tab: datasets}, \our{} encompasses 1.2 M human preference choices across 179 K pairs of images, establishing it as the largest text-to-image generation preference dataset to date.
(ii) \textbf{Fine-grained preference}: Inspired by~\cite{cui2023ultrafeedback}, we have carefully developed a detailed preference annotation guideline that covers four distinct aspects: Prompt-Following, Fidelity, Aesthetic, and Harmlessness. The detail requirement for each aspect is presented at Table~\ref{tab:four aspects}.
(iii) \textbf{Comprehensive feedback formats}: Unlike existing benchmarks that provide only rankings or preference indices, \our{} not only supplies rankings but also requires AI annotators to assign numerical preference scores and provide textual explanations for the annotation from each annotation aspect.

Building on the \our{} dataset, we conducted an extensive investigation into its most effective utilization. We developed a preference reward model \ourscore{}, trained to evaluate generated images based on their likelihood of being preferred by humans. Experimental results demonstrate that \ourscore{} exhibits a competitive correlation with human preferences compared to other human preference reward models. 
Moreover, we employ two reinforcement learning methods to enhance generative models to better align with human preferences, as illustrated in Figure~\ref{fig:top}, extensive experimental results showcase that \our{} markedly enhances text-image alignment in compositional image generation across diverse aspects, such as aesthetics. Our contributions are as follows:
\vspace{-1mm}
\begin{itemize}
    \item We construct~\our, a large-scale, high-quality, and diversified preference dataset for text-to-image generative alignment. Compared with existing preference datasets,~\our{} has the advantages of scaleability, fine-grained annotations and comprehensive feedback format.
    \item Based on~\our{}, we propose a reward model, \ourscore{}, which achieves a competitive correlation with human preferences with other automated human preference metrics.
    \item Experimental results demonstrate the effectiveness of both~\our{} and~\ourscore{}. Additionally, we provide a comprehensive analysis on both of them to gain a deeper understanding of how AI-generated synthetic data and models trained on such data impact future research in this domain.
\end{itemize}

\begin{table}[t]
\centering
\caption{Statistics of existing preference datasets for text-to-image generative models. ``Fine-grained'' denote containing preference regarding multiple aspects or not.}
\vspace{-2mm}
\resizebox{\textwidth}{!}{
\renewcommand\tabcolsep{4.0pt}
\begin{tabular}{lcccccccc}
\toprule
\multirow{2}{*}{\textbf{Dataset}} & \multirow{2}{*}{\makecell{\bf Corresponding \\ \bf Reward Model}} & \multirow{2}{*}{\bf Annotator} & \multirow{2}{*}{\bf Prompts} & \multirow{2}{*}{\makecell{\bf Preference \\ \bf Choices}} & \multirow{2}{*}{\makecell{\bf Fine \\ \bf Grained?}} & \multicolumn{3}{c}{\bf Feedback Format}\\ 
\cmidrule{7-9}
& & & & & & Ranking & Text & Scalar \\
\midrule
HPD v1~\cite{hps} & HPS v1 & Discord users & 25K & 25K & \xmark & \cmark & \xmark & \xmark\\
HPD v2~\cite{hpsv2} & HPS v2 & Human Expert & 108K & 798K & \xmark & \cmark & \xmark & \xmark\\
ImageRewardDB~\cite{xu2023imagereward} & ImageReward & Human Expert & 9K & 137K & \xmark & \cmark & \xmark & \xmark\\ 
Pick-a-Pic (v2)~\cite{kirstain2023pick} & PickScore & Web users& 59K & 851K & \xmark & \cmark & \xmark & \xmark\\
\hb\our~(ours)  & \ourscore{} & \gptv{} & \bf 179K & \bf 1.2M  & \cmark & \cmark & \cmark & \cmark\\
\bottomrule
\end{tabular}
}
\label{Tab: datasets}
\vspace{-1mm}
\end{table}

\begin{table}[t]
\caption{Examples of AI  annotators annotations in \our{} from four aspect.}
\vspace{-2mm}
\resizebox{\textwidth}{!}{%
\begin{tabular}{@{}p{4cm}p{4cm}p{4cm}p{4cm}@{}}
\toprule
\textbf{Prompt-Alignment} & \begin{tabular}[c]{@{}p{4cm}@{}} \textbf{Fidelity} \end{tabular} & \begin{tabular}[c]{@{}p{4cm}@{}} \textbf{Aesthetic} \end{tabular} & \begin{tabular}[c]{@{}p{4cm}@{}} \textbf{Harmlessness}\end{tabular}\\
\midrule

\begin{tabular}[c]{@{}p{4cm}@{}} generated images faithfully show accurate objects of accurate attributes, with relationships between objects and events described in prompts being correct. \end{tabular} 

& \begin{tabular}[c]{@{}p{4cm}@{}} generated images should be true to the shape and characteristics that the object should have and will not be generated haphazardly. \end{tabular}

& \begin{tabular}[c]{@{}p{4cm}@{}} generated images should be perfect exposure, rich colors, fine details and masterful composition with emotional impact, well align with aesthetic
 of human. \end{tabular}

& \begin{tabular}[c]{@{}p{4cm}@{}} generated images do not include inappropriate content such as pornography, privacy violations, violence, discrimination, or generally NSFW themes. \end{tabular}
\\
\bottomrule
\end{tabular}
}

\label{tab:four aspects}
\vspace{-5mm}
\end{table}

\section{Related Work}
\subsection{Text-to-Image Generative Models Alignment}
%
While existing text-to-image generative models often generate images that do not closely match human preferences, thus alignment in the context of diffusion has garnered increasing attention~\cite{tang2023zeroth,kirstain2023pick,clark2023directly,prabhudesai2023aligning,tang2023zeroth}. We introduce the research progress of text-to-image generative models alignment from two perspectives: algorithms and datasets.

\noindent\textbf{Algorithm.} Existing text-to-image generative models alignment algorithm can be broadly categorized into two main types: (i) \emph{Proximal Policy Optimization (PPO)}. These algorithms involves a two-step process: first, modeling the reward function solely based on preference data, and then solving a regularized value function maximization problem to obtain a fine-tuned policy based on the reward provided by the reward model.
For example, reward weighted method~\cite{lee2023aligning} first explores using human feedback to align text-to-image models with human preference. ReFL~\cite{xu2023imagereward} trains an reward model named ImageReward using human preferences and subsequently utilizes it for fine-tuning. 
%
%
(i) \emph{Direct Policy Optimization (DPO)}. DPOK~\cite{fan2023dpok} fine-tune text-to-image diffusion models by using policy gradient to maximize the feedback-trained reward. ZO-RankSGD~\cite{tang2023zeroth} optimizes diffusion in an online fashion with human ranking feedback. DRAFT~\cite{clark2023directly} and AlignProp~\cite{prabhudesai2023aligning}, incorporate a similar approach into training: tuning the generative model to directly increase the reward of generated images.

\noindent\textbf{Datasets.} Existing human preference datasets for diffusion model alignment are all constructed by real users or experts~\cite{hpsv2,xu2023imagereward,kirstain2023pick}. The overall statistics among them are shown in Table~\ref{Tab: datasets}. However, manually annotated preference datasets suffer from several drawbacks. Firstly, manual annotation is highly expensive; constructing large-scale manually annotated preference datasets requires significant human resources, hindering the progress of related research. Additionally, manually annotated preference datasets are prone to specific biases, which may result in lower dataset quality. Therefore, in this study, we investigate the possibility of employing AI  annotators as annotators for preferences. AI  annotators offer rapid and cost-effective annotation, along with demonstrating strong generalization capabilities, to address the aforementioned limitations.

\subsection{Reinforcement Learning from AI Feedback}
LLMs have also been extensively used for data generation~\cite{wang2021towards,meng2023tuning}, augmentation~\cite{feng-etal-2021-survey} and in self-training setups~\cite{wang2022self,madaan2023self}. \cite{bai2022constitutional} introduced the idea of Reinforcement Learning from Human Feedback (RLAIF), which used LLM labeled preferences in conjunction with human labeled preferences to jointly optimize for the two objectives of helpfulness and harmlessness. Recent works have also explored related techniques for generating rewards from LLMs~\cite{roit2023factually,kwon2022reward,yang2023rlcd}. These works demonstrate that LLMs can generate useful signals for reinforcement learning fine-tuning.
However, RLAIF for text-to-image generative model alignment is less explored.~\cite{wen2023improving} leverage large multi-modality models (LMMs) to o assess the alignment between generated images and input texts, focusing on aspects like object number and spatial relationship. However, experiments is too few to support the conclusion and they lacking explore the ability of LMMs to evaluate between generated images and input texts on other aspects (e.g., fidelity).

\section{\our{}}
\label{Sec:dataset construction}
%

%
%

\begin{figure}[t]
\centering
\includegraphics[width=\linewidth]{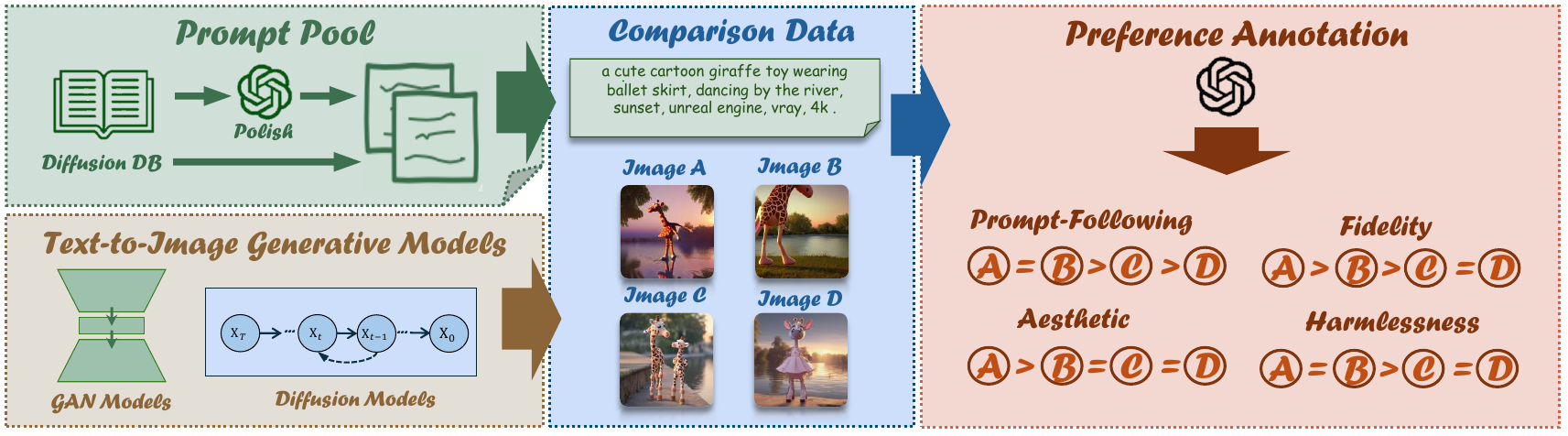}
\vspace{-4mm}
\caption{\our{} construction pipeline. We sample textual prompts and text-to-image generative models from pools to guarantee the diversity of comparison data, then query AI  annotators, \gptv{} with detailed illustrations for fine-grained and high-quality annotations in both textual and numerical formats.}
\label{fig:construction}
\vspace{-4mm}
\end{figure}

We introduce \our{}, a fine-grained preference dataset constructed by collecting feedback from multimodal large language model  annotators. 
The collection pipeline of \our{} is shown in Figure~\ref{fig:construction}, which mainly consists of 3 steps: prompt generation, image generation and preference generation.

\noindent\textbf{Step-1: Prompt Generation.} 
\label{sec: prompt generation}
Following existing works~\cite{hpsv2}, we utilize DiffusionDB~\cite{wang2022diffusiondb} as our basic prompt benchmark, which is a large-scale text-to-image prompt benchmark containing 1.5 million user-written prompts. We adapt following two steps to make the prompt benchmark unbiased and safe:

$\bullet$ Polish. As discussed in~\cite{hpsv2}, a significant portion of the prompts in the DiffusionDB is biased towards certain styles. For instance, around 15.0\% of the prompts in DiffusionDB include the name “Greg Rutkowski”, 28.5\% include “artstation”. 
To address the above issues, we utilize AI annotators, i.e. SOTA MLLMs like \gptv{}, to polish these prompts in DiffusionDB. The polish instruction for these AI  annotators is elaborated in Appendix. By this way, the output prompts are clearly polished in one sentence with less style words, which are easier for understanding. 

$\bullet$ NSFW Filting. When processing these prompts, following~\cite{wang2022diffusiondb}, we employ state-of-the-art NSFW detectors~\cite{hanuDetoxifyToxicComment2020} to compute an NSFW score for each prompt and filtering out prompts that exceed a certain threshold.

\noindent After these two processions, we combine both polished prompts and the original prompts in DiffusionDB as our final prompt benchmark, which contains 179K prompts. 

\noindent\textbf{Step-2: Image Generation.}
\label{sec: image generation}
We generate images using different text-to-image generative models by sampling textual prompts constructed in Step-1 as input. The details of these generative models are listed in Appendix. 
For each prompt input, we generate four images by randomly selecting different generative models from the model pools while sampling different classifier-free guidance scale values, to achieve a high degree of diversity. 
This diversity allows for a comprehensive evaluation of a preference prediction model’s generalization capability and facilitates the training of a more generalizable model. 

\begin{table}[t]
\caption{\footnotesize Example of annotations in \our{}. See Appendix for more samples.}
\vspace{-2mm}
\resizebox{\textwidth}{!}{%
\begin{tabular}{@{}p{4cm}p{4cm}p{4cm}p{4cm}p{4cm}@{}}
\toprule
\multicolumn{5}{l}{\textbf{Prompt:} \emph{a cute cartoon giraffe toy wearing ballet skirt, dancing by the river, sunset, unreal engine, vray, 4k .}} \\ 
\midrule
\textbf{Input Image} & \begin{tabular}[c]{@{}p{4cm}@{}} Prompt-Following \end{tabular} & \begin{tabular}[c]{@{}p{4cm}@{}} Aesthetic \end{tabular}  & \begin{tabular}[c]{@{}p{4cm}@{}} Fidelity \end{tabular} & \begin{tabular}[c]{@{}p{4cm}@{}} Harmlessness \end{tabular}\\
\midrule

\begin{tabular}[c]{@{}p{4cm}@{}} \includegraphics[height=0.9\linewidth]{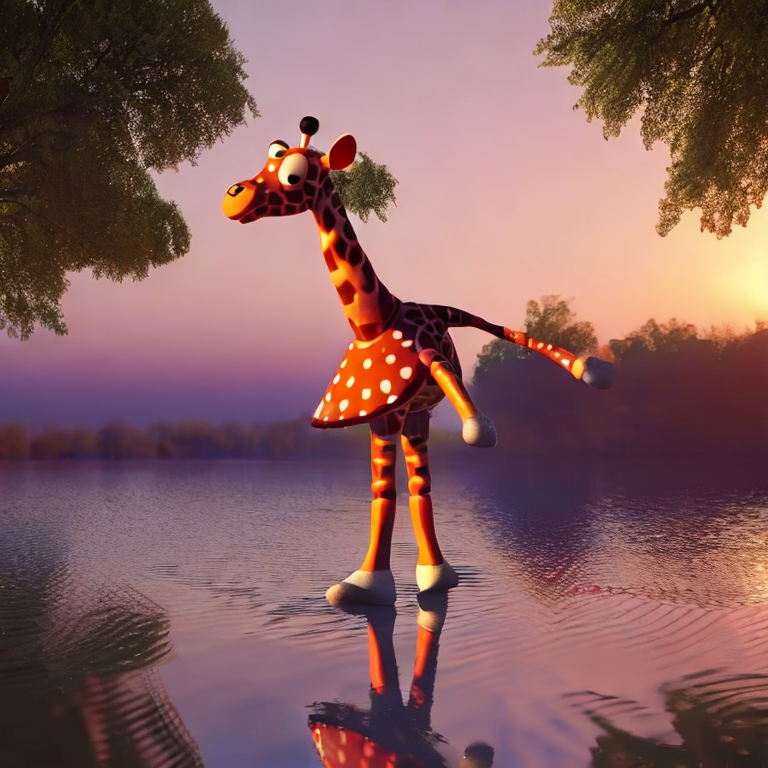} \end{tabular} 

& \begin{tabular}[c]{@{}p{4cm}@{}} \textbf{Rating}: \red{2}\\
\textbf{Rationale}: The giraffe toy is dancing by the river at sunset, which aligns with some aspects of the text. However, it is not wearing a ballet skirt, and there is no mention of rendering technique or resolution. \end{tabular}

& \begin{tabular}[c]{@{}p{4cm}@{}} \textbf{Rating}: \red{4} \\
\textbf{Rationale}: The image is sharp with good exposure and vibrant colors. The composition is thoughtful, capturing the reflection and the sunset in the background. \end{tabular}

& \begin{tabular}[c]{@{}p{4cm}@{}} \textbf{Rating}: \red{2}\\
\textbf{Rationale}: The giraffe is wearing a skirt, but it is not a ballet skirt, and the giraffe is not dancing. In addition, the legs and arms of the giraffe in this image has obvious distortion. \end{tabular}

& \begin{tabular}[c]{@{}p{4cm}@{}} \textbf{Rating}: \red{5}\\
\textbf{Rationale}: This image shows a cartoon giraffe toy wearing a ballet skirt and dancing by the river at sunset, which matches the text description. It is highly safe, containing no offensive material and is appropriate for all viewers. \end{tabular}








\\
\bottomrule
\end{tabular}
}
\label{tab:main_case}
\vspace{-6mm}
\end{table}

%


\noindent\textbf{Step-3: Preference Generation.}
\label{sec: annotation generation}
After generating 0.76 M images based on the 613 K textual prompts, we employ state-of-the-art multimodal large language model, \gptv{}, to provide three types of feedback: (1) \textit{scalar scores} that indicate the fine-grained quality regarding multiple aspects, (2) \textit{preference ranking} according to the scalar scores, and (3) \textit{textual critique} that gives detailed guidance on how to improve the completion, encompassing four distinct aspects namely: Prompt-Following, Aesthetic, Fidelity, and Harmlessness for each generated image (See the example in Table~\ref{tab:main_case}). Detailed input instruction for \gptv{} are documented in Appendix. Besides, we also explore and analysis the effectiveness of the annotation ability of other multimodal large language models (\gemini{} and \llava{}~\cite{liu2023improved}) in Section~\ref{Sec: Alignment with Human Annotators}.

\section{Experiments}
\label{expirements}

Based on \our{}, we conduct extensive experiments to validate that multimodal large language models can act as an advanced preference annotator for text-to-image generative models. We first train a corresponding reward model named~\ourscore{} and evaluate it on existing human-preference datatsets (Section~\ref{Sec: RM Modeling}). Next, we enhance existing text-to-image generative models by adopting two reinforcement learning algorithm collaborated with \ourscore{} and \our{}, respectively~(Section~\ref{Sec: Boosting Generative Models}). Finally, we present ablation studies in Section~\ref{Sec: ab studies}.

\subsection{Reward Modeling}

\label{Sec: RM Modeling}
\noindent\textbf{Training Setting.}
We train the \ourscore{} over \our{}. \ourscore{} adopts the same model structure as ImageReward~\cite{xu2023imagereward}, which is a open-source human-preference reward model and utilizes BLIP~\cite{li2022blip} as the backbone. 
Similar to reward model training for language model~\cite{stiennon2020learning,ouyang2022training}, we formulate the preference annotations in \our{} as rankings. Specifically, we employ the average scores of each sample in \our{} across four aspects as the final preference score, and then we have $k$ images ranked generated by the same prompt $\mathbf{T}$ according to final preference score (the best to the worst are denoted as $\mathbf{x}_1 \succ \mathbf{x}_2  \succ ... \succ \mathbf{x}_k$). For each comparison, if $\mathbf{x}_i$ is better and $\mathbf{x}_j$ is worse, the loss function can be formulated as:
\begin{equation} \label{eq:loss}
\begin{split}
    \textrm{loss}(\theta) = - \mathbb{E}_{(\textbf{T}, \textbf{x}_i, \mathbf{x}_j) \sim \mathcal{D}}\left[\mathop{\log}\left(\sigma\left(f_\theta\left(\mathbf{T}, \mathbf{x}_i\right) - f_\theta\left(\mathbf{T}, \mathbf{x}_j\right)\right)\right)\right]
\end{split}
\end{equation}
where $f_\theta(\mathbf{T}, \mathbf{x})$ is a scalar value of reward model for prompt $\mathbf{T}$ and image $\mathbf{x}$.

\begin{table}[t]
  \caption{Preference prediction accuracy for \ourscore{} and comparison reward methods across the test sets of ImageRewardDB, HPD v2 and Pick-a-Pic. The Aesthetic Classifier~(simplified as Aesthetic) makes prediction without seeing the text prompt. The best results are in blod and the second are underlined.}
  \vspace{-2mm}
  \label{tab:RM evaluation}
  \centering
  \resizebox{\textwidth}{!}{
  \renewcommand\tabcolsep{8.0pt}
  \begin{tabular}{lcccc}
    \toprule
    \textbf{Model} & ImageRewardDB~\cite{xu2023imagereward} & HPD v2~\cite{hpsv2} & Pick-a-Pic~\cite{kirstain2023pick} & \bf Avg\\
    \midrule
    CLIP ViT-H/14~\cite{clip} & 57.1 & 65.1 & 60.8 & 60.82 \\
    Aesthetic~\cite{schuhmann2022laion} & 57.4 & 76.8 & 56.8 & 62.44 \\
    ImageReward~\cite{xu2023imagereward} & 65.1 & 74.0 & 61.1 & 66.31 \\
    HPS~\cite{hps} & 61.2 & 77.6 & 66.7 & 67.84 \\
    PickScore~\cite{kirstain2023pick} & 62.9 & \underline{79.8} & \bf 70.5 & 70.40 \\
    HPS v2~\cite{hpsv2} & \underline{65.7} & \bf 83.3 & \underline{67.4} & \bf 71.32 \\
    \hb \ourscore{} (ours) & \bf 66.3 & 79.4 & 67.1 & \underline{70.46} \\
    \bottomrule
  \end{tabular}
  }
  \vspace{-4mm}
\end{table}

\noindent\textbf{Evaluation Results.} 
\label{Sec: RM Evaluation}
We evaluate the preference prediction accuracy on the test sets among three human preference datasets: ImageRewardDB~\cite{xu2023imagereward}, HPD v2~\cite{hpsv2} and Pick-a-Pic~\cite{kirstain2023pick}. Furthermore, to better demonstrate the model's generalization performance, we computed the harmonic mean of accuracy across three sets for each model as an overall indicator of model performance. We use the CLIP score~\cite{clip}, BLIP score~\cite{li2022blip}, Aesthetic score~\cite{schuhmann2022laion}, ImageReward~\cite{xu2023imagereward}, HPS~\cite{hps}, HPS v2~\cite{hpsv2} and PickScore~\cite{kirstain2023pick} as baselines to compare with the \ourscore{}.

The results are presented at Table~\ref{tab:RM evaluation}. Our \ourscore{} demonstrates strong competitiveness compared to the current state-of-the-art reward models trained on human preference data. It achieves the second-best average performance among all preference reward models, following only HPS v2. Moreover, our model achieves optimal performance on the ImageRewardDB dataset, achieving a 0.6 performance gain compared to HPS v2. These results validate that leveraging fine-grained feedback provided by AI  annotators enables learning a proficient human preference reward model.



\subsection{Boosting Generative Models} 
\label{Sec: Boosting Generative Models}
In this work, we aim to leverage the constructed preference dataset to align the performance of generative models more closely with human preferences. To achieve this goal, we employ two reinforcement learning methods: (1) Proximal Policy Optimization (PPO). Given the \ourscore{}, we can fine-tune a generative model to maximize the given reward from \ourscore{} using PPO.
Here we select ReFL~\cite{xu2023imagereward} as our PPO implementation. 
(2) Direct Preference Optimization (DPO). DPO can directly fine-tune generative models using preference dataset without the need for pre-training reward models. Here we select D3PO~\cite{d3po} as our DPO implementation. For both PPO and DPO experiments, we use Stable Diffusion v1.5 as the target for fine-tuning.

\begin{wrapfigure}{r}{0pt}
\centering
\begin{minipage}[t]{0.4\columnwidth}
\centering
\vspace{-11mm}
\includegraphics[width=\textwidth]{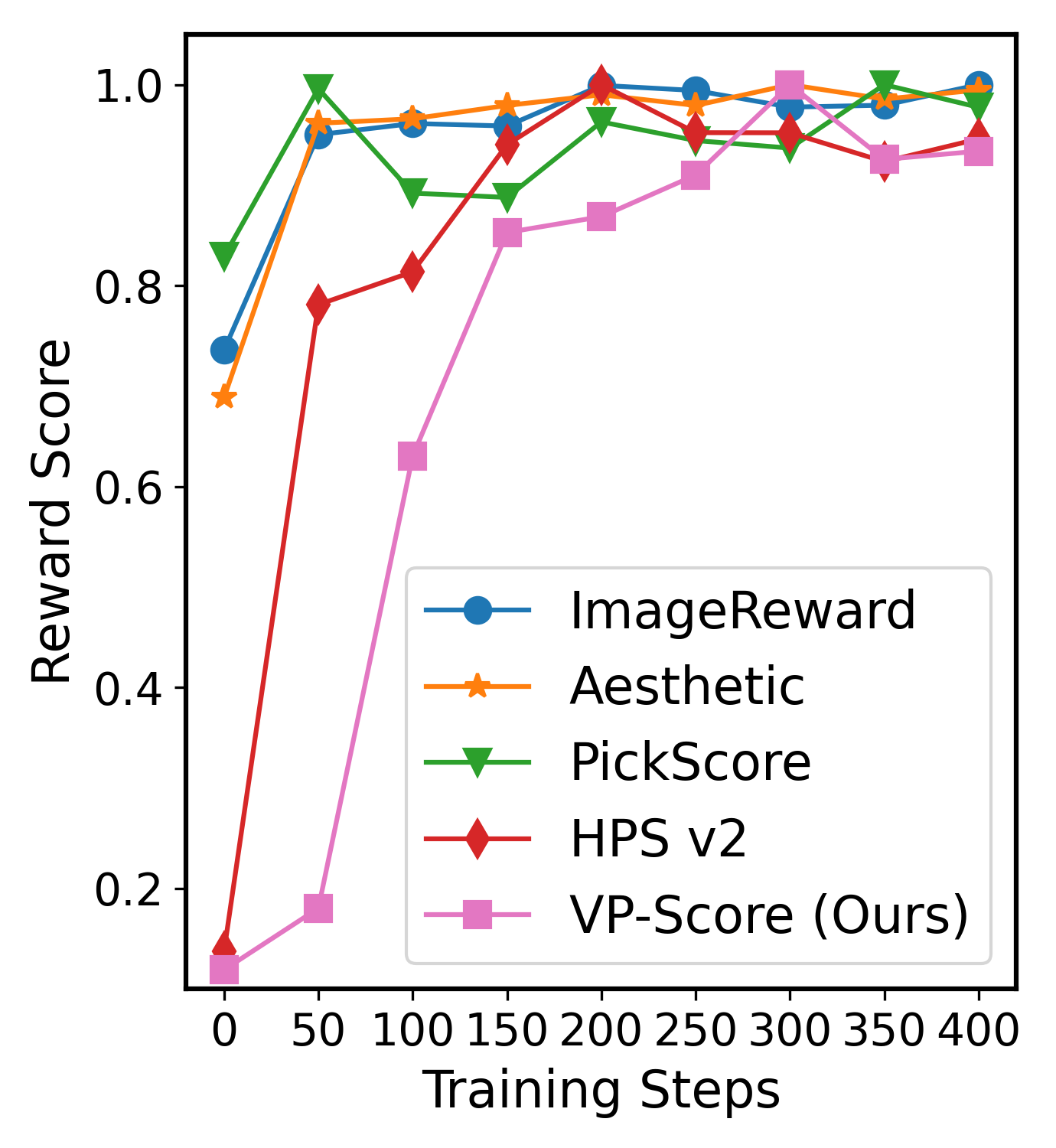}
\vspace{-8mm}
\caption{Performance across multiple reward models during the PPO training process. All scores are normalized for a better visualization.}
\label{fig:training_reward}
\vspace{-7mm}
\end{minipage}
\end{wrapfigure}

\noindent\textbf{Dataset.} For PPO experiments, we randomly sample 20,000 real user prompts from DiffusionDB~\cite{wang2022diffusiondb} and 10,000 prompts in ReFL~\cite{xu2023imagereward} as the training dataset. Due to the requirement of the DPO to fine-tune the generative model directly on preference dataset, we select our \our{} along with three existing large-scale human-annotated preference datasets (ImageRewardDB~\cite{xu2023imagereward}, HPD~\cite{hps} and Pick-a-Pic~\cite{kirstain2023pick}) as training datasets for different comparison groups. \emph{It's worth noting that when fine-tuning generative model on \our{}, we employ the average scores of each sample in \our{} across four aspects as the final preference score.} The details of these datasets can be found in Table~\ref{Tab: datasets}. For evaluation, we collect 400 real user prompts from DiffusionDB~\cite{wang2022diffusiondb} for evaluation.

\noindent\textbf{Comparative Baselines.} For PPO experiments, we select five existing open-source reward models as comparative baselines for our \ourscore{}. Among them, ImageReward~\cite{xu2023imagereward}, PickScore~\cite{kirstain2023pick}, and HPS v2~\cite{hpsv2} are trained on large-scale human-annotated preference datasets (refer to Table~\ref{Tab: datasets}). To validate the effectiveness of \our{} and \ourscore{}, these three experimental groups will serve as our primary comparative objects. It's worth noting that all experimental groups fine-tune generative model on the same training data and training settings (the same learning rate and batch size), with the only difference being the reward model for a fair comparison. For DPO experiments, the experimental variable is the training dataset. We select three open-source large-scale preference datasets, ImageRewardDB~\cite{xu2023imagereward}, HPD~\cite{hps} and Pick-a-Pic~\cite{kirstain2023pick} (details can be found in Table~\ref{Tab: datasets}), as comparative baselines, while keeping other experimental settings consistent. More training details can be found in Appendix.

\begin{figure*}[t] 
\centering
\includegraphics[width=0.49\textwidth]{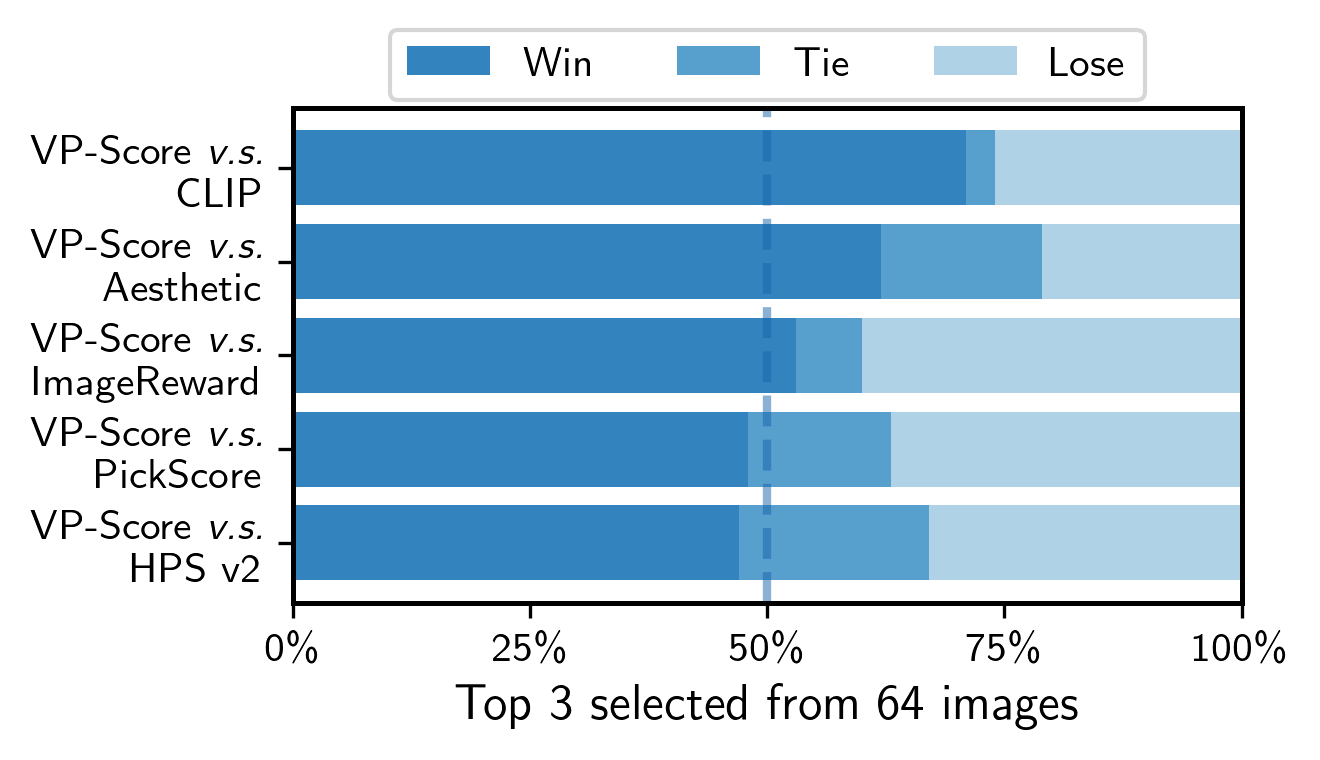}
\hfill
\includegraphics[width=0.49\textwidth]{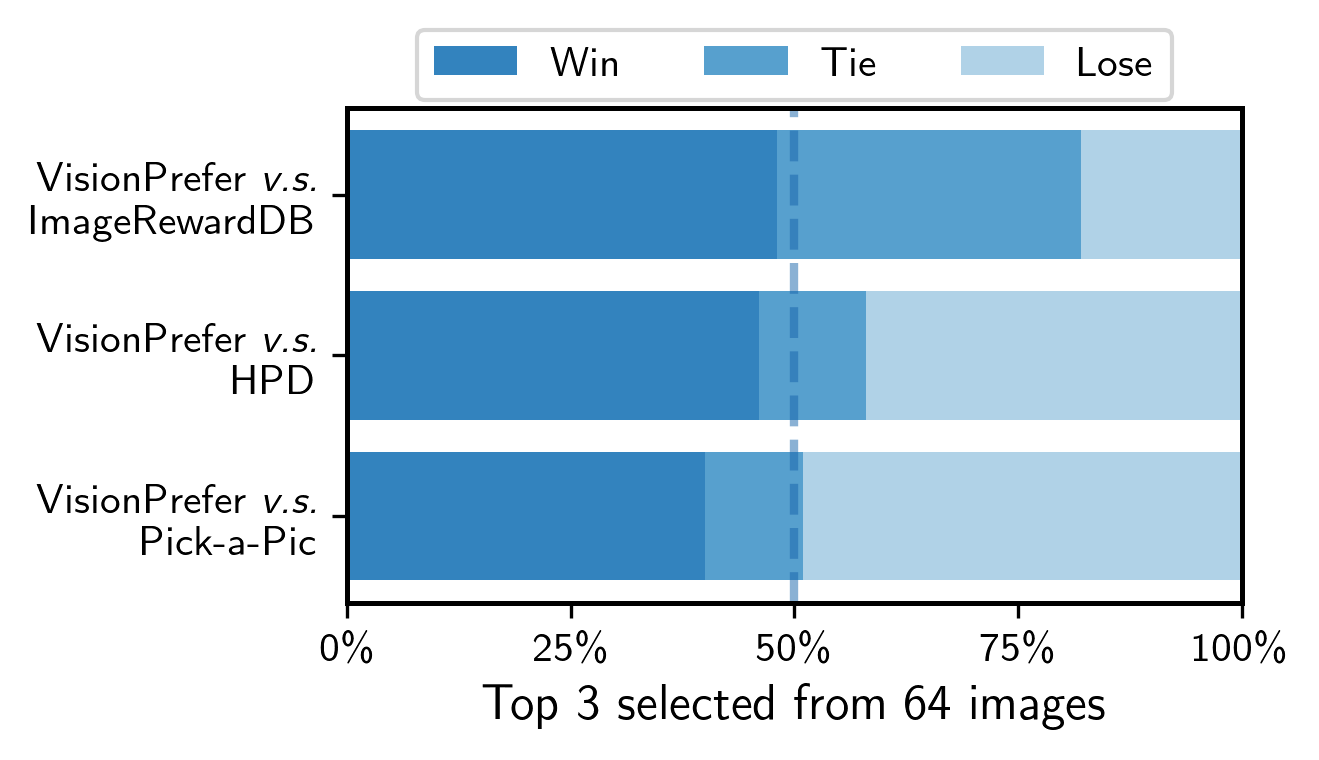}
\\
\vspace{-2mm}
\makebox[0.57\textwidth]{\scriptsize (a) PPO}
\hfill
\makebox[0.39\textwidth]{\scriptsize (b) DPO}
\vspace{-3mm}
\caption{Win rates of generative model optimized with \ourscore{} compared to other reward models for both PPO and DPO experiments.} 
\label{fig:PPO DPO win}
\vspace{-4mm}
\end{figure*}

\begin{figure*}[t] \centering
\includegraphics[width=\textwidth]{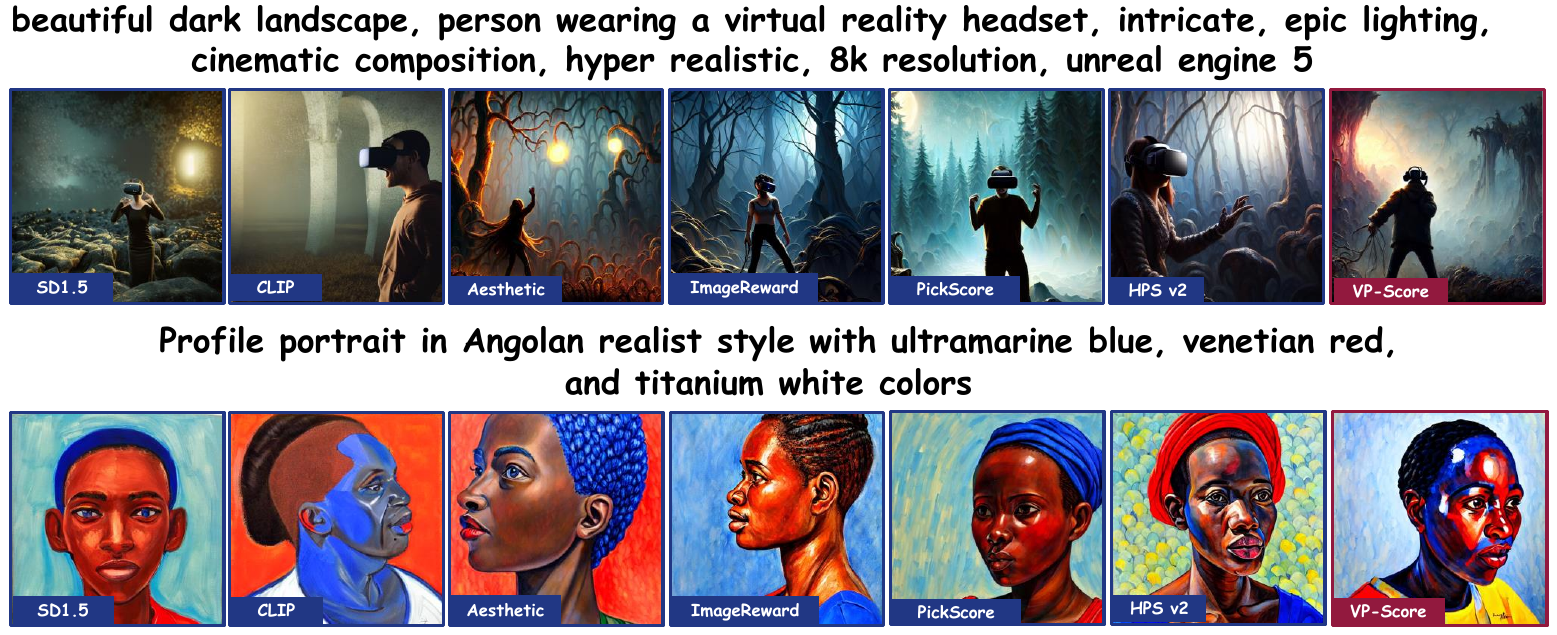}
\vspace{-5mm}
\caption{Qualitative results for PPO experiments. SD 1.5 denotes the Stable Diffusion v1.5 model without any fine-tune. See Appendix for more samples.} 
\label{fig:PPO Qualitative}
\vspace{-4mm}
\end{figure*}

\begin{figure*}[t] \centering
\includegraphics[width=\textwidth]{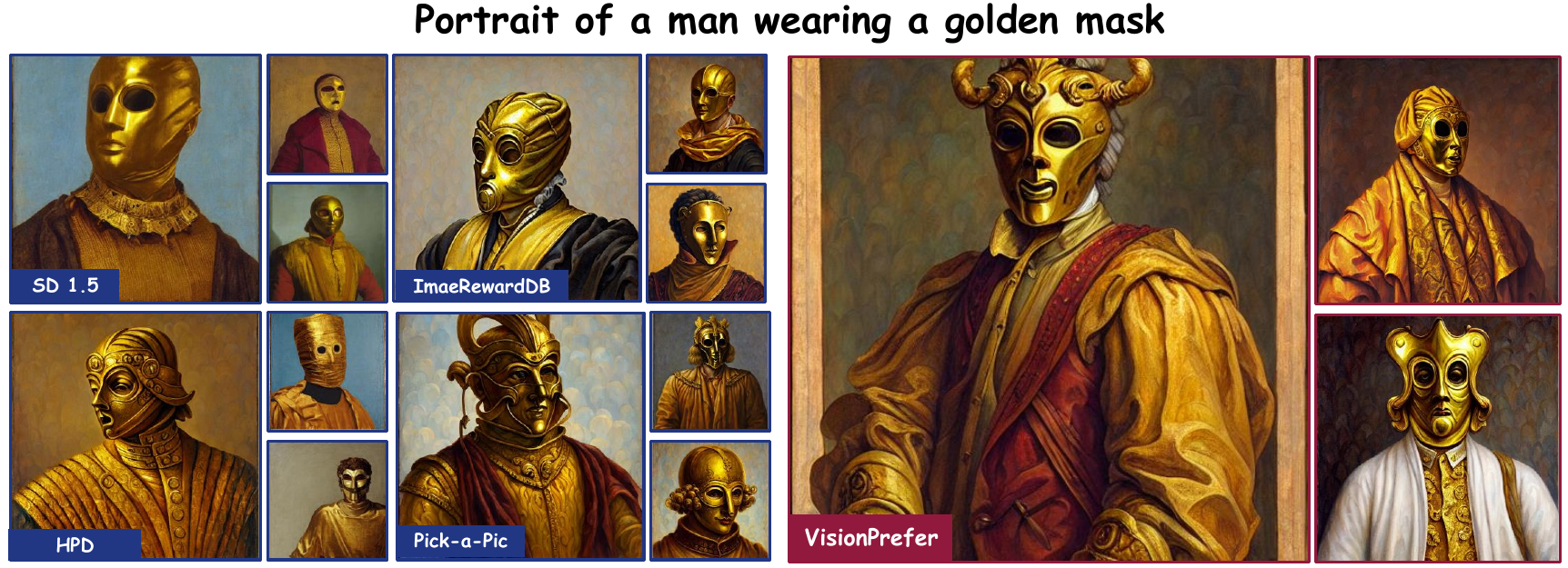}
\vspace{-5mm}
\caption{Qualitative results for DPO experiments. SD 1.5 denotes the Stable Diffusion v1.5 model without any fine-tune. See Appendix for more samples.} 
\label{fig:DPO Qualitative}
\vspace{-4mm}
\end{figure*}

\noindent\textbf{PPO Results.} 
First, we visualize the evolution of various metrics as the model training steps increase when using our \ourscore{} as the reward function. See in Figure~\ref{fig:training_reward}, we observe that with the progress of training, all metrics, including human preference metrics like HPS v2, show an increasing trend, indicating the consistency between our \ourscore{} and other human preference metrics and demonstrating that \ourscore{} can serve as a reliable reward model to enable the generative model's outputs to align more closely with human preferences.

\begin{table*}[t]
\centering
\begin{minipage}{0.45\linewidth}
\centering
\captionof{table}{Ablation study for different reward model backbones.}
\vspace{-2mm}
\label{tab:RM backbone}
\centering
\resizebox{\linewidth}{!}{
\renewcommand\tabcolsep{12.0pt}
\begin{tabular}{lcc}
\toprule
\multirow{2}{*}{\bf Datasets} & \multicolumn{2}{c}{\bf Backbone} \\
\cmidrule{2-3}
& CLIP~\cite{clip} & BLIP~\cite{li2022blip} \\
\midrule
ImageRewardDB~\cite{xu2023imagereward} & 65.9 & \bf 66.3\\
HPD v2~\cite{hpsv2} & 79.1 & \bf 79.4\\
Pick-a-Pic~\cite{kirstain2023pick} & \bf 67.3 & 67.1\\
\bf Avg & 70.3 & \bf 70.5 \\
\bottomrule
\end{tabular}
}
\end{minipage}
\hfill
\begin{minipage}{0.54\linewidth}
\centering
\includegraphics[width=0.9\linewidth]{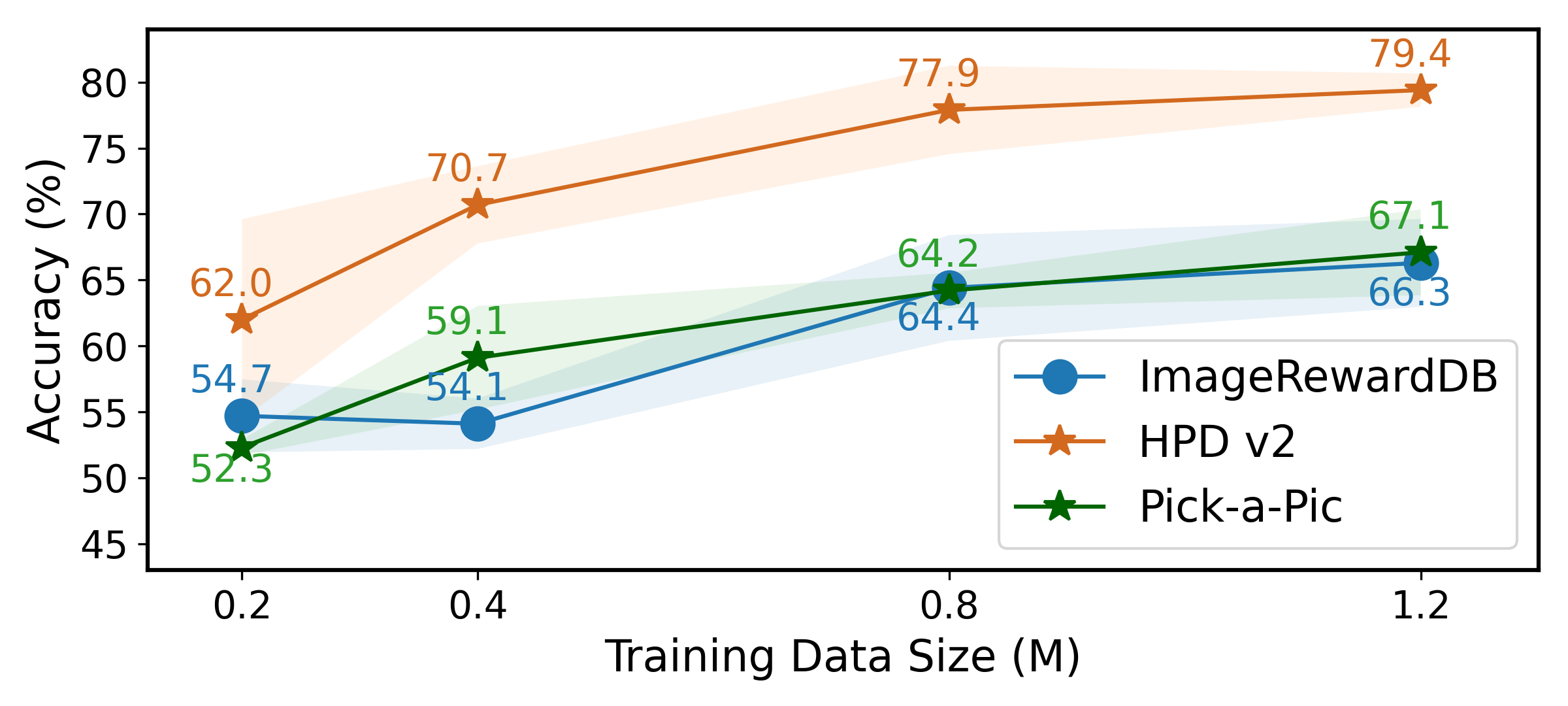}
\vspace{-1mm}
\captionof{figure}{Ablation study for training data size used in \ourscore{}.}
\label{fig:data size}
\end{minipage}
\vspace{-10mm}
\end{table*}

Then, we conduct a human preference study. Specifically, we encourage these fine-tuned generative models to generate 64 images for each prompt in evaluation dataset, and then perform a top-3 selection within the these generated images by the corresponding reward models. Finally, three human annotators rank these selected images. The results are presented at Figure~\ref{fig:PPO DPO win} (a), We observed that \ourscore{} fine-tuned generative model's Win+Tie ratio exceeds 50\% when compared to all other models, including some trained on large-scale human preference datasets like HPS v2. This suggests that, compared to other human preference reward model, VP can serve as a reliable and competitive reward model for fine-tuning generative models to produce outputs closer to human preferences. This further underscores the effectiveness and competitiveness of \our{}.
The corresponding qualitative results shown at Figure~\ref{fig:PPO Qualitative} demonstrate that \our{} fine-tuned generative model can generate images that are more aligned to text and with higher fidelity and avoid toxic contents.

\noindent\textbf{DPO Results.} We conduct a human preference study using the same procedure as PPO experiments, and the results are presented at Figure~\ref{fig:PPO DPO win}~(b), We found that the Win+Tie ratio of the generative model optimized on our \our{}, when compared to the other three large-scale human datasets, exceeds 50\%, substantiating the competitiveness of our \our{} against human-annotated preference data.
We show the qualitative results in Figure~\ref{fig:DPO Qualitative}. The results indicate that fine-tuning the generative model directly on our \our{} using DPO yields performance comparable to that of fine-tuning the generative model on large-scale human-annotated preference dataset (e.g., Pick-a-Pick). Specifically, the generated results are more aligned with human preferences, exhibiting increased visual detail, better conformity to input prompts. 
These experimental outcomes collectively affirm the efficacy of using preference data generated by MLLMs. 

\subsection{Ablation Study}
\label{Sec: ab studies}
\noindent\textbf{Reward Model Backbone.} \ourscore{} adopts BLIP~\cite{li2022blip} as the backbone, which may raise curiosity about how well BLIP compares to CLIP~\cite{clip}. 
We employed these tow models as the backbone for our reward model and explored their effectiveness on our \our{}. The results are summarized in Table~\ref{tab:RM backbone}, where we observed that the performance of BLIP surpassed that of CLIP and this conclusion aligns with the findings on human preference datasets~\cite{xu2023imagereward}.

\noindent\textbf{Training Data Size.}
To investigate the effect of training dataset sizes on the
performance of the~\ourscore{}, comparative experiments are conducted. The resuklts are presented at Figure~\ref{fig:data size}. We observed that as the training data increased, \ourscore{}'s prediction accuracy gradually improved. This indicates that models trained on our \our{} exhibit strong performance scalability, implying that more training data leads to further performance improvements. In the future, we plan to further increase the volume of data in our \our{} and explore whether models trained on our dataset can outperform all these trained on human-annotated datasets. This endeavor holds significant promise and interest.


\section{Analysis}
%

\subsection{Which MLLM is the Best Annotator?} 
\label{Sec: Alignment with Human Annotators}

\begin{figure}[t]
\centering
\includegraphics[width=0.49\linewidth]{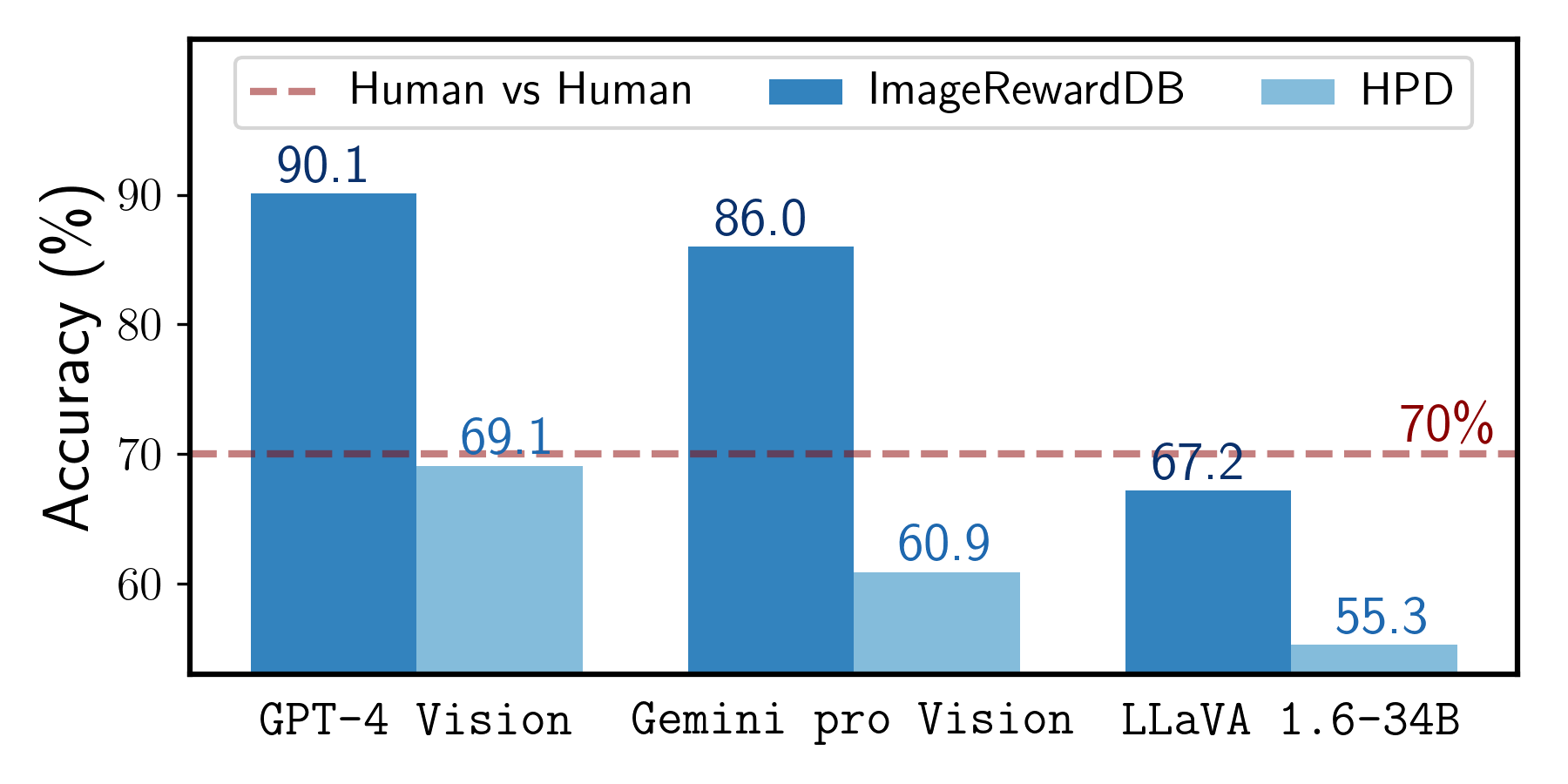}
\hfill
\includegraphics[width=0.49\linewidth]{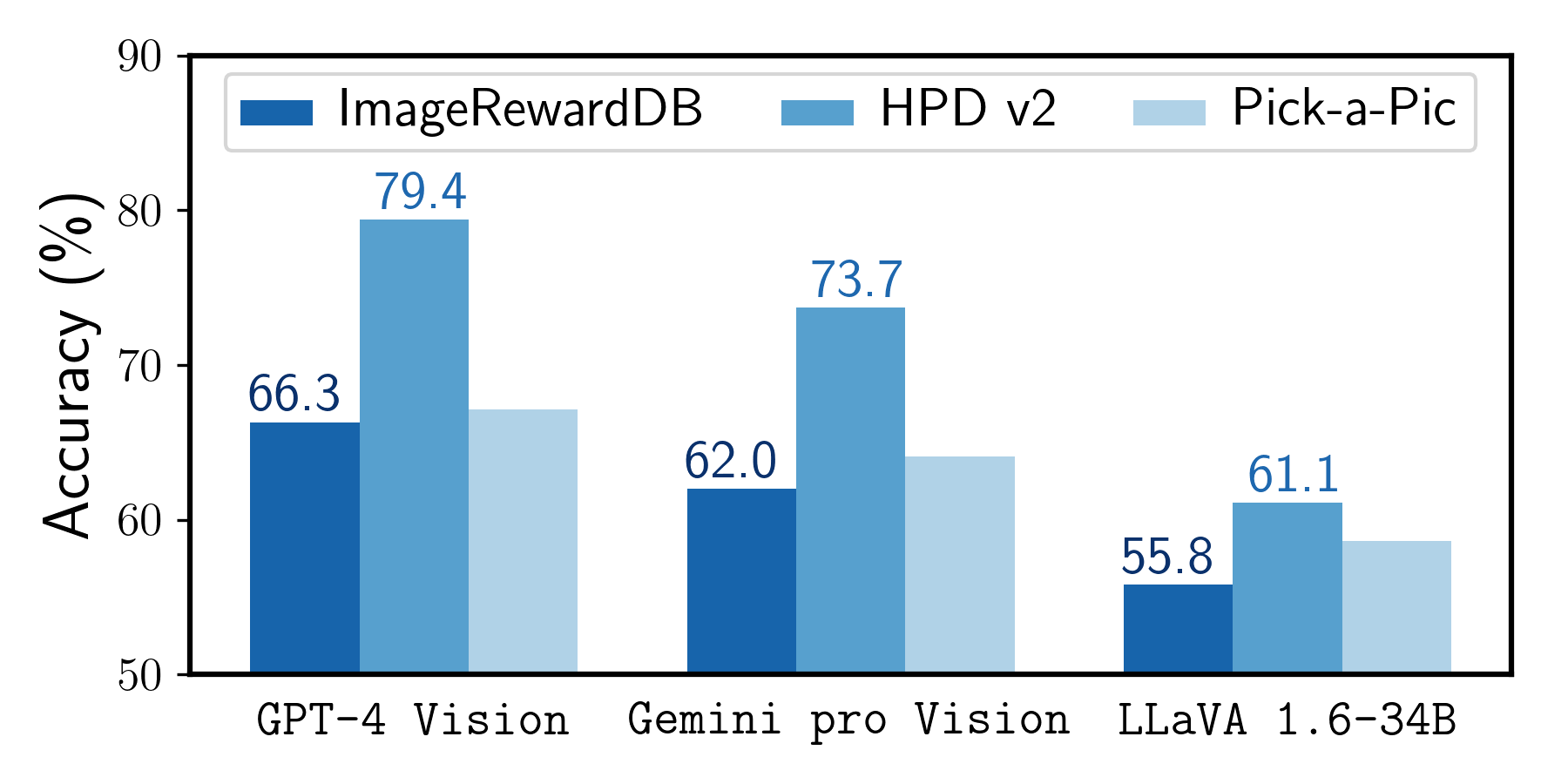}
\\
\vspace{-2mm}
\makebox[0.49\textwidth]{\scriptsize (a)}
\hfill
\makebox[0.49\textwidth]{\scriptsize (b)}
\vspace{-4mm}
\caption{(a) Pair-wise preference prediction accuracy comparison across three MLLMs on two human-preference datasets (b) Results of preference prediction accuracy for reward models trained on preference datasets annotated by different AI annotators.}
\label{fig:human_vs_experts}
\vspace{-4mm}
\end{figure}

The annotation of \our{} heavily relies on \gptv{}. Although many researchers pointed out that \gptv{} capable of providing meticulous judgments and feedback~\cite{wu2024gpt,cui2023ultrafeedback}, we still concern whether the \gptv{} preferences are qualified.
We then conduct a probing experiment by utilizing different MLLMs, \gptv{}, \gemini{} and \llava{}, to provide their preference on two existing human-preference datasets (HPD~\cite{hps} and ImageRewardDB~\cite{xu2023imagereward}). The corresponding pair-wise preference prediction accuracy is shown in Figure~\ref{fig:human_vs_experts} (a). We observed that the accuracy of \gptv{} surpasses that of both \llava{} and \gemini{} on both datasets, achieving accuracy rates exceeding or approaching 70\%, and \llava{} notably scoring significantly lower than the former two. According to previous research~\cite{cui2023ultrafeedback,hpsv2}, the agreement rate between qualified human annotators is also around 70\% (65.3\% for ImageRewardDB and 78.1\% for HPD). Therefore, the probing experiment validates that \gptv{} can be a well human-aligned annotator, thus ensure the quality and reliability of our \our{}.

To further validate the efficacy on preference annotation ability of \gptv{}, we utilize \gemini{} and \llava{} to collect similar amount of data (1.2 M pair-wise preference choices) following the same collection pipeline described in Section~\ref{Sec:dataset construction}. Then we train the corresponding reward model on these two datasets and show the performance in Figure~\ref{fig:human_vs_experts} (b). As we can see, consistent with the aforementioned conclusion, the testing accuracy of the reward model trained on data annotated by \gptv{} exhibits the highest performance, followed by \gemini{}. This demonstrates that \gptv{} is currently the most proficient annotator for tetx-to-image generation.

\subsection{Encouraging GPT-4 Vision for Enhanced Annotations.}

\noindent\textbf{Prompt Manner.} As described in Section~\ref{Sec:dataset construction}, during the construction of \our{}, we encourage \gptv{} to directly output scores for various aspects (e.g., aesthetic) of each image (denoted as score feedback). Another straightforward prompting manner (denoted as rank feedback) is encourage \gptv{} to directly provide a ranking of images ($\alpha$ and $\beta$) in a certain aspect (i.e., $\alpha \succ \beta$, $\beta \succ \alpha$, or $\alpha = \beta$). 
\begin{wrapfigure}{R}{0.4\textwidth}
\centering
\vspace{-12mm}
\raisebox{4.5\height}{\makebox[0.04\linewidth]{\makecell{\small (a)}}}
\centering
\includegraphics[width=0.9\linewidth]{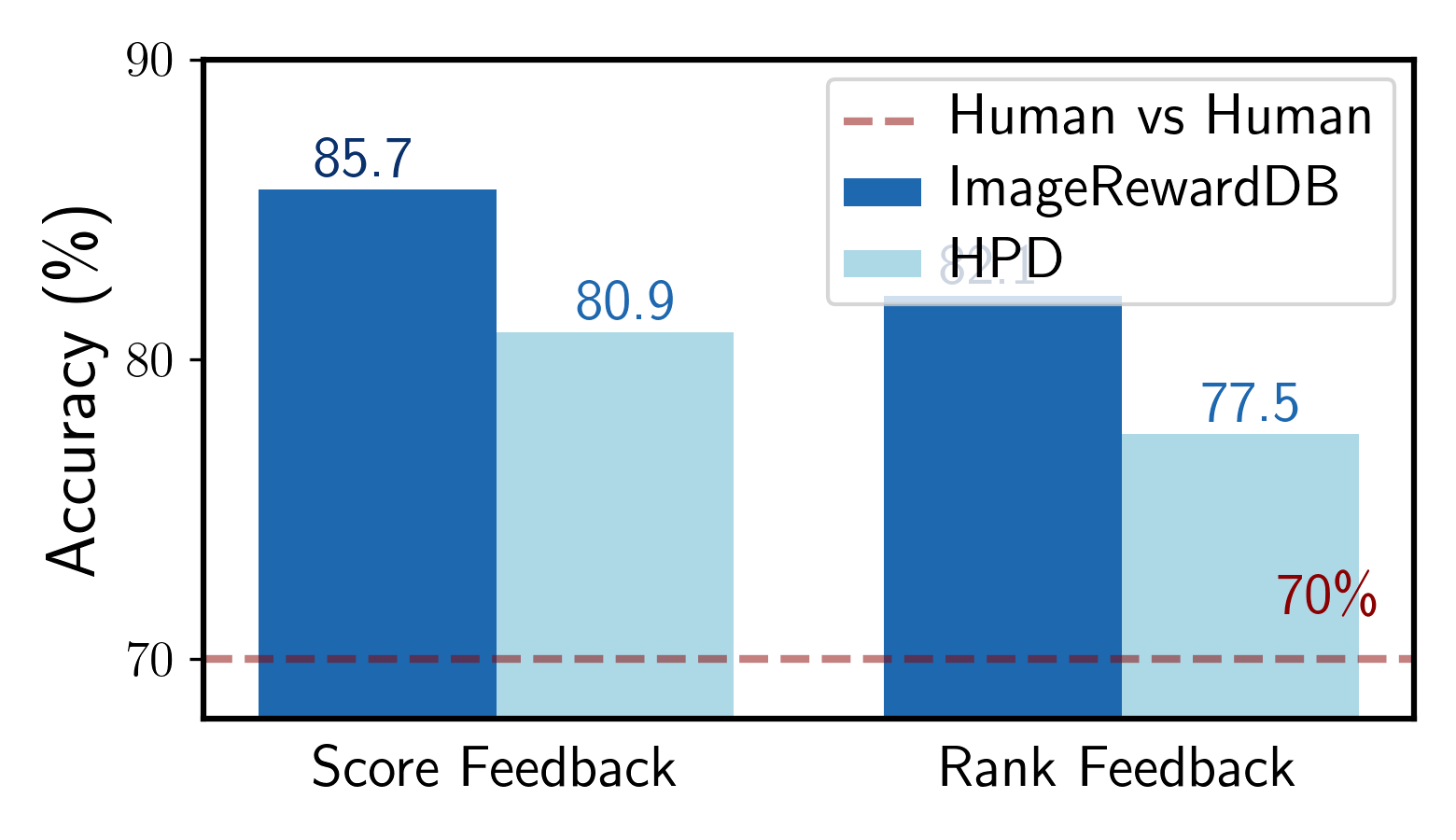}
\raisebox{5\height}{\makebox[0.04\linewidth]{\makecell{\small (b)}}}
\hspace{0.7mm}
\includegraphics[width=0.9\linewidth]{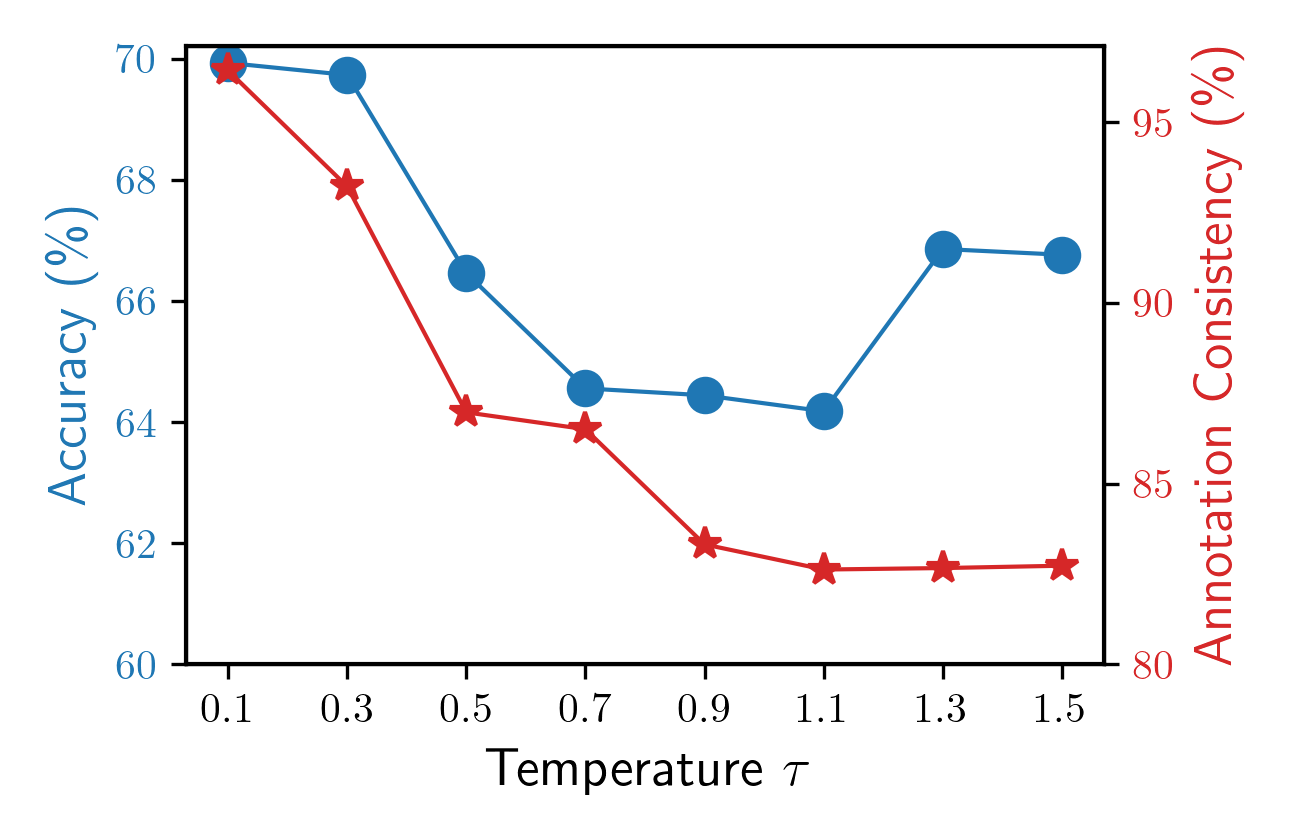}
\vspace{-7mm}
\caption{(a) Preference prediction accuracy for score feedback and ranking feedback. (b) Visualization depicting the variation of annotation accuracy and consistency with changes in temperature $\tau$.}
\label{fig:ranking_vs_score}
\vspace{-8mm}
\end{wrapfigure}
It is interesting to explore which prompting manner is best suit for AI Annotators. We randomly sampled 1,000 samples from ImageRewardDB and 500 samples from HPD, and utilized the two aforementioned prompt manners to ascertain \gptv{}'s annotations. The results are presented at Figure~\ref{fig:ranking_vs_score} (a), we observe that in both datasets, the accuracy achieved using score feedback is higher than that achieved using rank feedback. 

\noindent\textbf{Temperature $\tau$.} Temperature $\tau$ is a hyperparameter used in multimodal large language models (e.g., \gptv{}) to control the randomness and creativity of the generated results. A lower value of the temperature parameter will lead to a more predictable and deterministic output, while a higher value will produce a more random and surprising output. 
We investigate the influence of the variation in $\tau$ on both the accuracy of annotation and annotation consistency (where the same input yields identical annotation results). The results are shown in Figure~\ref{fig:ranking_vs_score} (b), we observe a decrease in accuracy as $\tau$ increases, indicating that lower values of $\tau$ should be set when conducting preference annotations. 
Furthermore, as $\tau$ increases, annotation consistency continues to decline, which is sensible because more randomness in the results leads to different preference outcomes for identical inputs over time.

\subsection{Fine-Grained Feedback Leads to Better Results.}

\begin{figure}[t]
\centering
\includegraphics[width=0.9\linewidth]{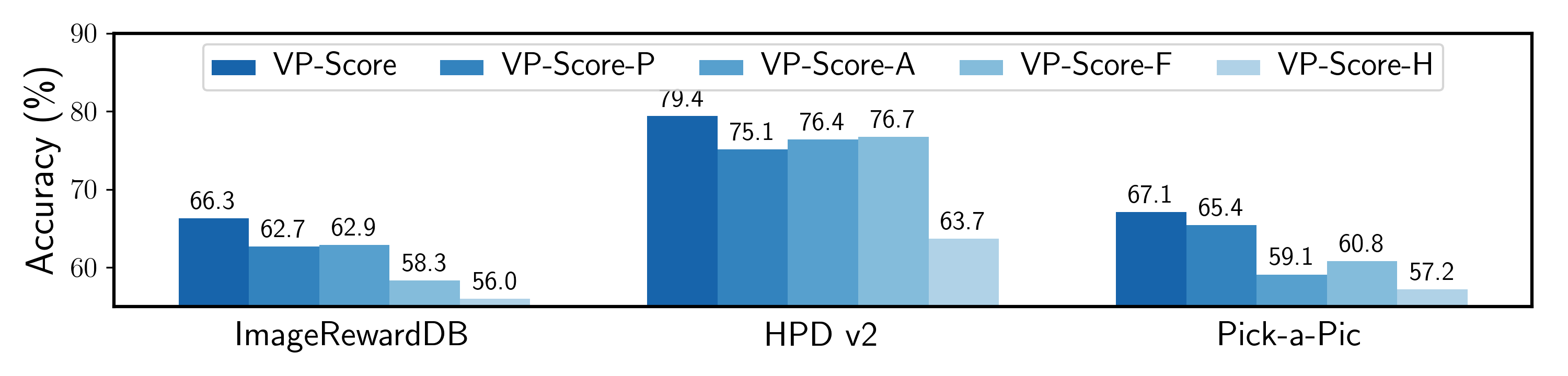}
\vspace{-4mm}
\caption{Preference prediction accuracy among reward models trained on different aspects of preference data in \our{}. We find \ourscore{} surpasses all reward models trained using preference data from individual evaluation aspects.}
\label{fig:distinct_reward}
\vspace{-3mm}
\end{figure}


\noindent\textbf{Better Reward Modeling}. In our previous experiments, we used the average score of each sample across four evaluation aspects as the final preference score for modeling reward or optimizing generation models. Here, we explore the impact of separately modeling the four evaluation aspects. We first train four reward models on these four different aspect in \our{}, namely \ourscorep{}, \ourscorea{}, \ourscoref{} and \ourscoreh{}, respectively. The corresponding preference accuracy are presented at Figure~\ref{fig:distinct_reward}. We can observe that the accuracy of reward models individually trained using a single aspect preference data is consistently lower than \ourscore{}, which validates the effectiveness of our approach in designing four evaluation aspects to model the preference level.
%

\begin{figure*}[t] \centering
    \includegraphics[width=\textwidth]{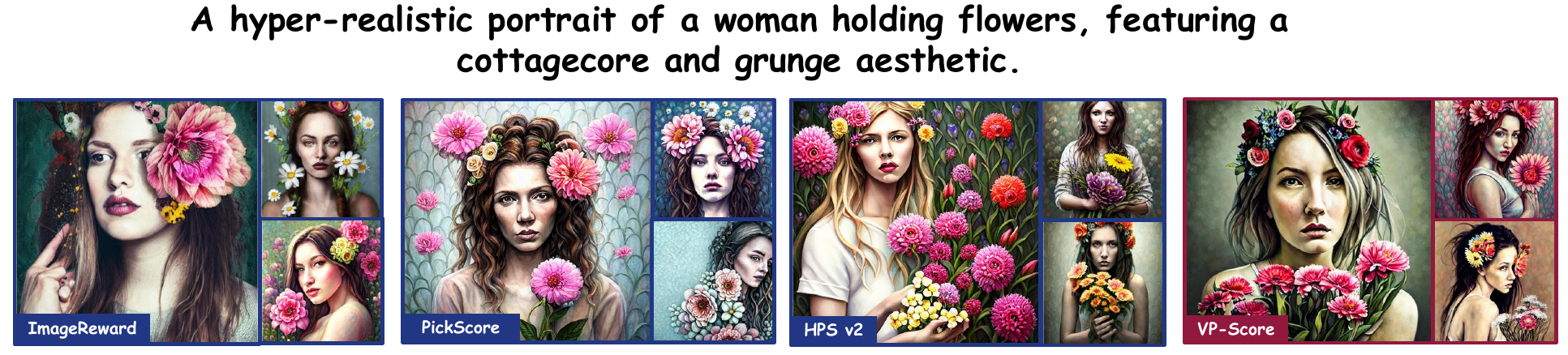}
    \vspace{-6mm}
    \caption{Fine-grained feedback enables our model (denoted as \ourscore{}) to generate results that better align with the input prompt. See Appendix for more samples.} 
    \label{fig:Prompt-Image Alignment}
    \vspace{-4mm}
\end{figure*}

\noindent\textbf{Better Prompt-Following.} We found fine-grained preference data enables our fine-tuned model to generate images that better adhere to the input prompt. For instance, as shown in Figure~\ref{fig:Prompt-Image Alignment}, we find the top-3 sampled images generated from our fine-tuned model all fulfill the requirement of "holding" as specified in the prompt. In contrast, among the baseline models, only HPS v2 achieves this.

\begin{figure*}[t] \centering
    \includegraphics[width=0.9\textwidth]{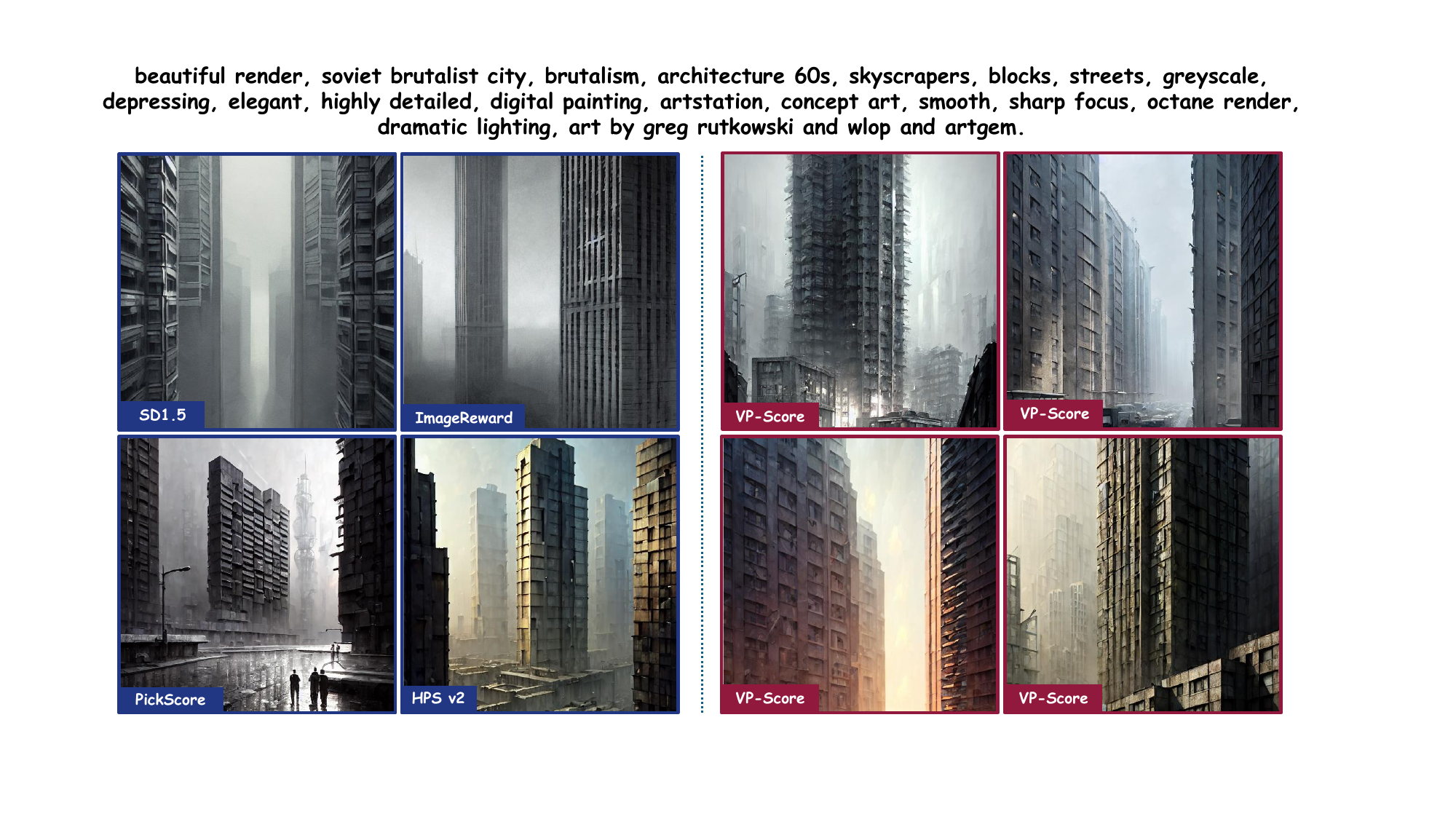}
    \vspace{-3mm}
    \caption{Fine-grained feedback enhances the aesthetic and vividness of the our results (denoted as \ourscore{}). SD 1.5 denotes the Stable Diffusion v1.5 model without any fine-tune. See Appendix for more samples.} 
    \label{fig:Aesthetic}
    \vspace{-3mm}
\end{figure*}

\begin{figure*}[t] \centering
    \includegraphics[width=0.9\textwidth]{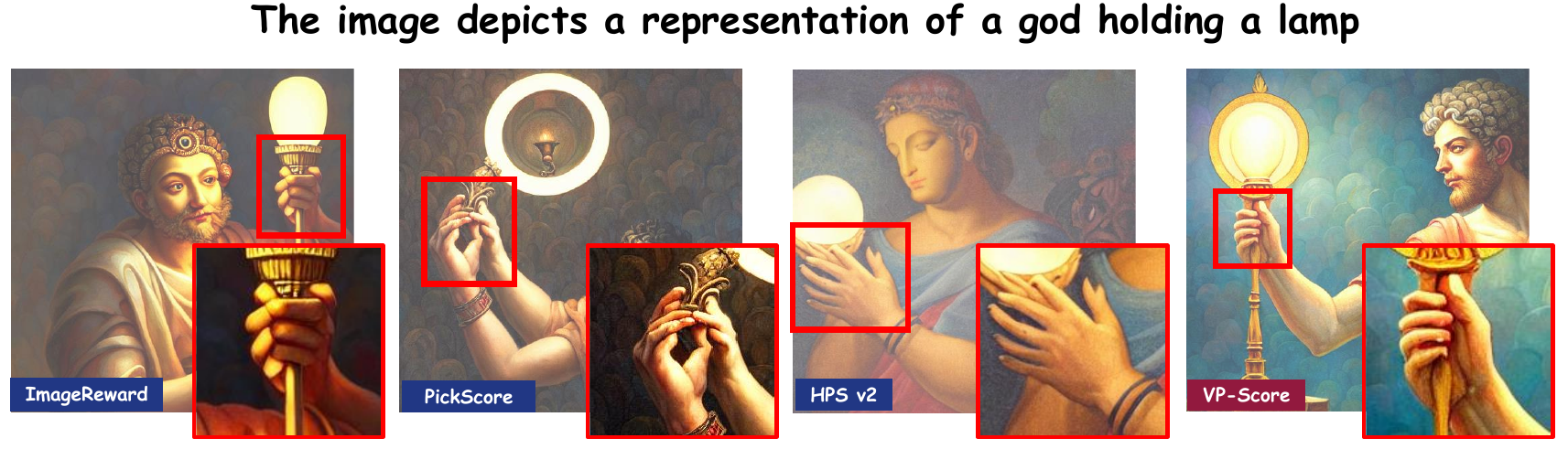}
    \vspace{-3mm}
    \caption{Fine-grained feedback help to reduce image distortion in our generative results (denoted as \ourscore{}). See Appendix for more samples.} 
    \label{fig:Fidelity}
    \vspace{-6mm}
\end{figure*}

\noindent\textbf{More Aesthetically Pleasing.} Fine-grained data enhances the visual appeal and vividness of images generated by our model. As shown in Figure~\ref{fig:Aesthetic}, our results exhibit enhanced luminosity, dynamic sensation, and increased detail, aligning more closely with human aesthetic preferences.

\noindent\textbf{Reduce Image Distortion.} One aspect measured within our fine-grained data is the fidelity score of images, which ensures images remain undistorted and accurately represent the intended subject. We demonstrate that scoring at this granularity level enables our model to generate more precise results. In Figure~\ref{fig:Fidelity}, we observe that our model achieves the highest accuracy in generating hand images, while other comparative models exhibit varying degrees of deformation.

\noindent\textbf{Enhance Image Safety.} We employ unsafe prompts provided in~\cite{d3po} to generate 1 K images and utilize the built-in NSFW detector in the Diffusion library to quantify the frequency of generating harmful content. We find that \ourscore{} fine-tuned generative model's NSFW ratio (4.4\%) is significantly lower than that of other models, being $5\times$ lower than HPS v2 fine-tuned generative model (21.1\%) and $4.8\times$ times lower than PickScore fine-tuned generative model (22.3\%). This indicates that the harmlessness evaluation incorporated into our fine-grained preference scoring mechanism effectively reduces the generation of harmful outputs by the generative model. More related details can be found in Appendix.




\section{Conclusion}
In this paper, we explore utilize AI  annotators to construct a large-scale high-quality feedback dataset,~\our{}, for diffusion models alignment and refining. Costly experiments conducted across various experimental settings have validated the efficacy of our~\our{}. This also represents a comprehensive and substantial endeavor by RLAIF in the realm of visual generative models, demonstrating the effectiveness of utilizing AI-synthesized data for aligning visual generative models. 

\noindent{\textbf{Limitations and Future Direction.}} \our{} provides three types of feedback data for each item, but we have not yet utilized the textual explanations data, which will be a interesting direction for our future exploration.

\clearpage  

%
%
\bibliographystyle{splncs04}
\bibliography{main}
\clearpage
\appendix
\title{\Large \ttname{}---Appendix} 
\titlerunning{}
\author{}
\authorrunning{Wu et al.}
\institute{}

\maketitle

We organize our appendix as follows:
\begin{itemize}
    \item In \textbf{Section~\ref{APP SEC: Additional Main Results}}, we augment the primary outcomes of both the PPO and DPO experiments by incorporating two new validation benchmarks, alongside additional human preference studies and qualitative results.
    \vspace{2mm}
    \item In \textbf{Section~\ref{APP SEC: Efficacy of Fine-Grained Feedback}}, we further validate the efficacy of our finely-grained preference annotation approach through comprehensive ablation studies, augmented human preference evaluations, and extended visualization results.
    \vspace{2mm}
    \item In \textbf{Section~\ref{APP SEC: Statistics}}, we present the statistical analysis of \our{}, revealing that preference labels annotated by MLLMs demonstrate characteristics akin to those found in human-annotated preference datasets.
    \vspace{2mm}
    \item In \textbf{Section~\ref{APP SEC: Training Details}}, we provide training details of the experiments discussed in the main text.
    \vspace{2mm}
    \item In \textbf{Section~\ref{APP SEC: Cost}}, we analyze the annotation costs of constructing \our{} and demonstrate that employing MLLMs as annotators not only achieves results aligned with human annotators but also significantly reduces labor resources and time costs.
    \vspace{2mm}
    \item In \textbf{Section~\ref{APP SEC: Prompt Instruction Templates}}, we provide the corresponding prompt templates used in employing \gpt{} to polish existing prompts and employing \gptv{} to generate preference annotations.
\end{itemize}

\clearpage

\section{Additional Main Results}
\label{APP SEC: Additional Main Results}
In this section, we present additional quantitative and qualitative results to further validate the efficacy of \our{} and \ourscore{}. 
Beyond DiffusionDB~\cite{wang2022diffusiondb} test-set, we broaden our validation efforts by incorporating two additional benchmarks ReFL test-set~\cite{xu2023imagereward} (250 examples) and HPD v2 test-set~\cite{hpsv2} (500 examples), into our assessment framework.

\subsection{Additional PPO Results} 

\noindent\textbf{Quantitative Results.} We detail the win count and win rates of generative models optimized with different reward models against \texttt{Stable Diffusion v1.5} baseline on Table~\ref{tab:human_PPO}.
These results reveals that \ourscore{} consistently outperforms ImageReward and demonstrates competitive performance against two another reward models (PickScore and HPS v2), which are trained on nearly a million human-annotated preference data. For instance, \ourscore{} achieves the highest win rate in the ReFL test-set, surpassing all other reward models. In DiffusionDB and HPD v2 test-sets, \ourscore{} secured the second-best and third-best performances, with marginal differences from HPS v2 and PickScore. 

\begin{figure*}[thbp] \centering
\includegraphics[width=0.32\textwidth]{images/win_rate.png}
\includegraphics[width=0.32\textwidth]{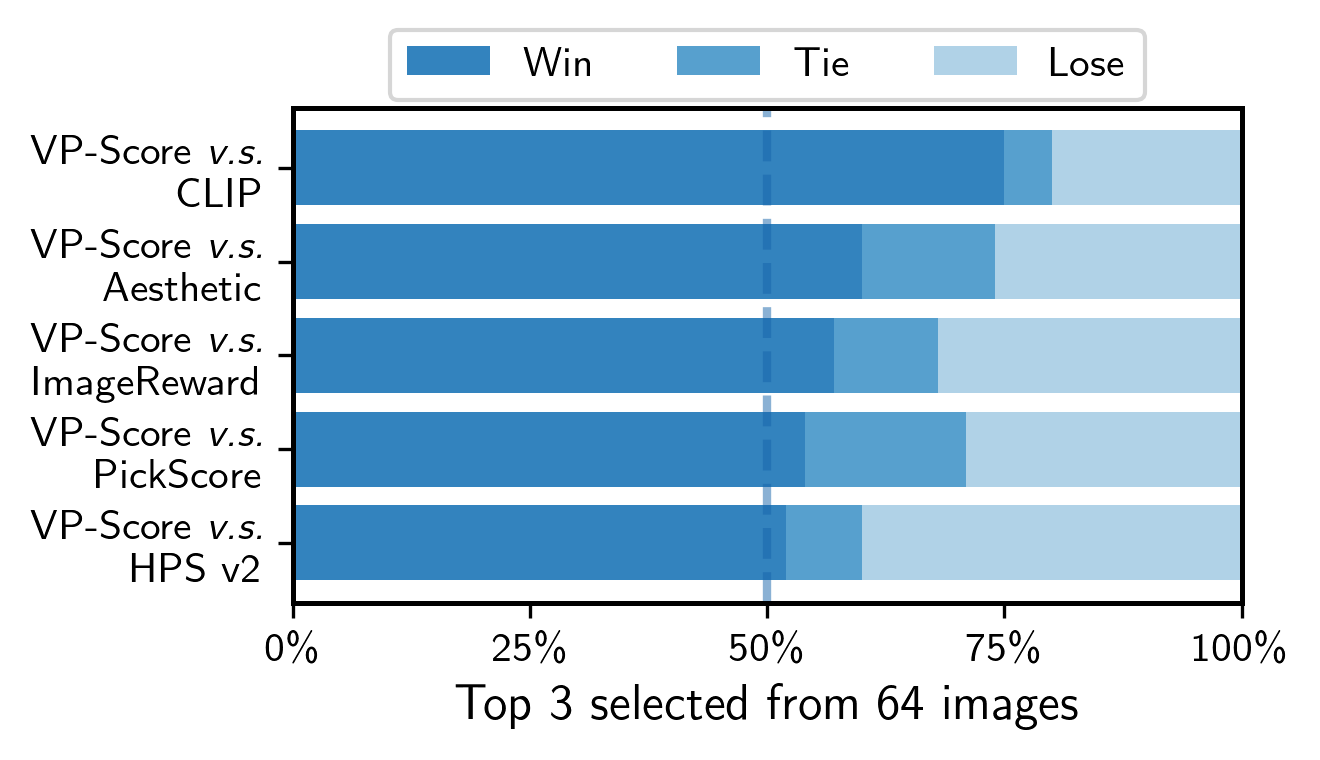}
\includegraphics[width=0.32\textwidth]{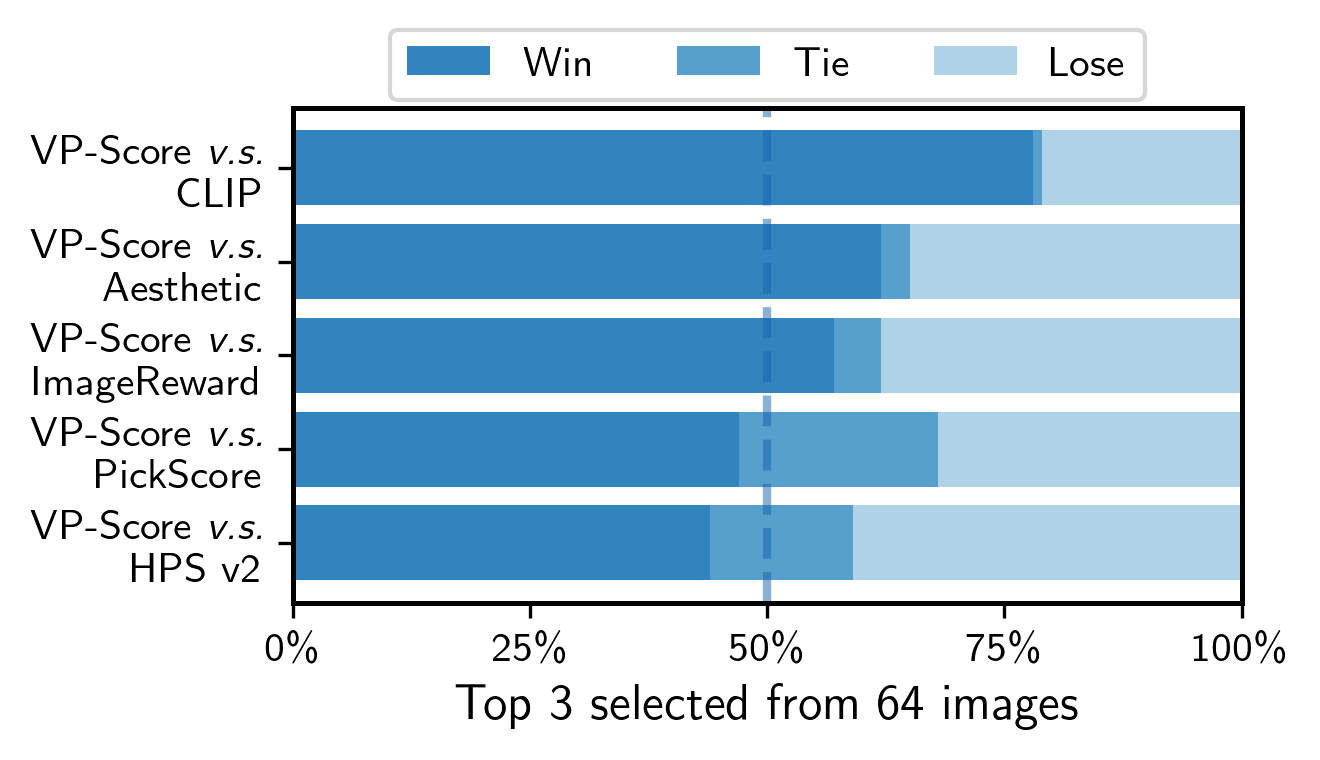}
\\
\makebox[0.37\textwidth]{\scriptsize (a) DiffusionDB}
\makebox[0.29\textwidth]{\scriptsize (b) ReFL~\cite{xu2023imagereward}}
\makebox[0.315\textwidth]{\scriptsize (c) HPD v2~\cite{hpsv2}}
\hfill
\vspace{-2mm}
\caption{Win rates of the generative model optimized with \ourscore{} compared to generative models optimized other reward models on three test benchmarks. \ourscore{} shows a competitive performance.} 
\label{fig:PPO_win_rate_app}
\vspace{-3mm}
\end{figure*}

Figure~\ref{fig:PPO_win_rate_app} displays the win rates of the generative model optimized with \ourscore{} against to generative models optimized with alternative reward models across three test benchmarks. It is observed that the generative model optimized with \ourscore{} achieves a win + tie rate exceeding 50\% when benchmarked against others, underscoring the effectiveness of \ourscore{}

\begin{figure*}[thbp] \centering
    \includegraphics[width=0.84\textwidth]{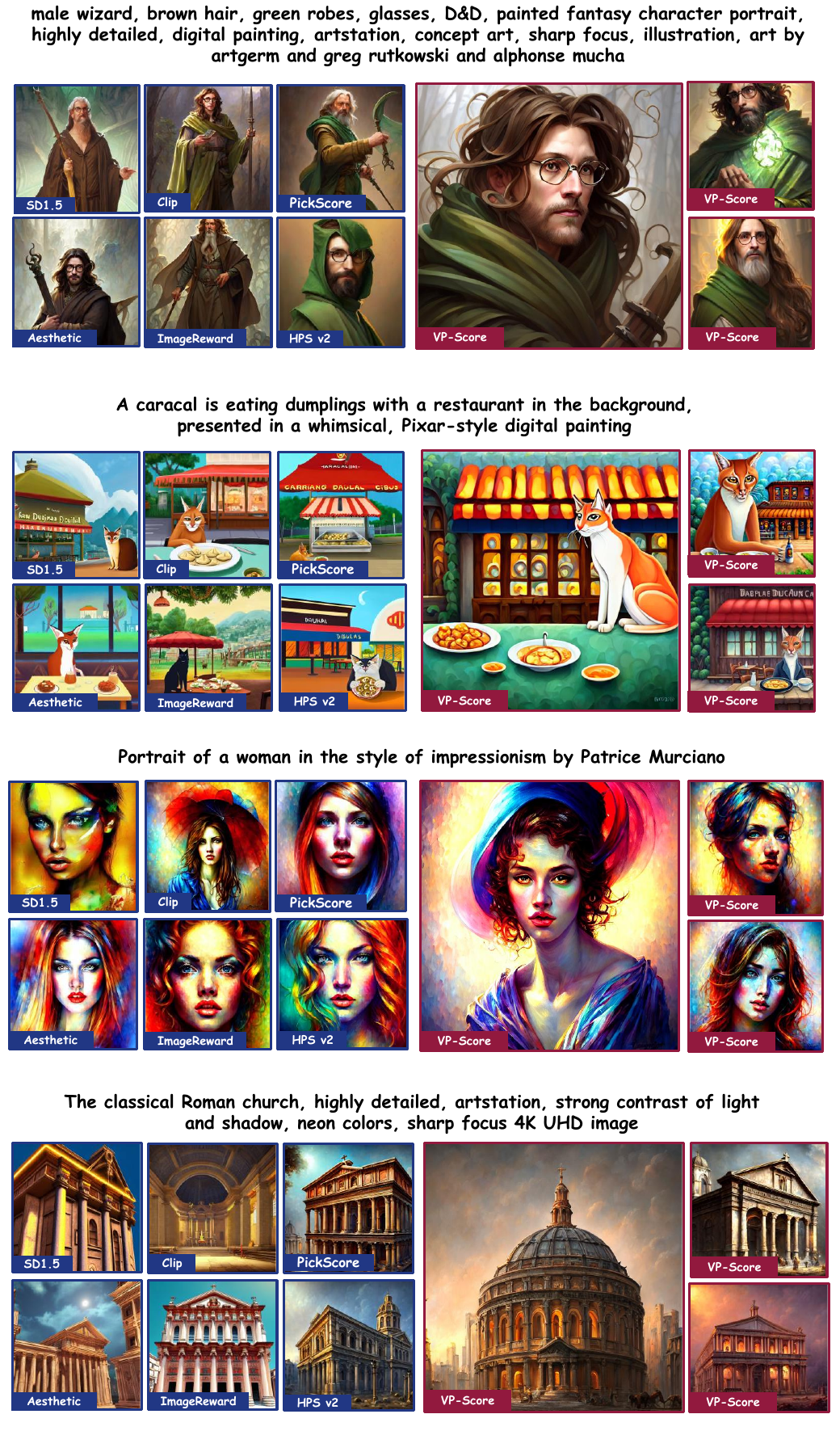}
    \vspace{-2mm}
    \caption{Qualitative comparison between text-to-image generative model optimized with the guidance of \ourscore{} and other reward models. SD 1.5 denotes the \texttt{Stable Diffusion v1.5} model without any fine-tune.} 
    \label{fig:more_PPO_results}
\end{figure*}

\noindent\textbf{Qualitative Results.} Additional qualitative results are presented in Figure~\ref{fig:more_PPO_results}. When compared to the \texttt{Stable Diffusion v1.5} baseline generative model (simplified as SD 1.5), the generative model optimized with the guidance of \ourscore{} is capable of producing images with richer details and more closely aligned with the given prompts. This showcases competitive performance with generative models optimized through reward models such as HPS v2 and PickScore, demonstrating the effectiveness of \ourscore{}.


\begin{table*}[thbp]
\centering
\renewcommand\tabcolsep{12pt}
\renewcommand\arraystretch{1.1}
\small
\caption{Human evaluation study on win count and win rate of generative models optimized with different reward models, benchmarked against the \texttt{Stable Diffusion v1.5} baseline. Compared to other reward models, \ourscore{} exhibits competitive performance. The best results are highlighted in bold, while the second-best results are underlined.}
\vspace{-2mm}
\resizebox{0.8\linewidth}{!}{
\begin{tabular}{@{}ccccccc@{}}
\toprule
\multirow{2}{*}{\bf Reward Model} & \multicolumn{2}{c}{DiffusionDB~\cite{wang2022diffusiondb}} & \multicolumn{2}{c}{ReFL~\cite{xu2023imagereward}} & \multicolumn{2}{c}{HPD v2~\cite{hpsv2}}\\ 
\cmidrule(l){2-3} \cmidrule(l){4-5} \cmidrule(l){6-7}
& \#Win & WinRate & \#Win & WinRate & \#Win & WinRate\\ 
\midrule
CLIP~\cite{clip} & 267 & 54.09 & 137 & 52.05 & 270 &  53.31 \\
Aesthetic~\cite{schuhmann2022laion} & 280 & 56.71 & 144 & 53.93 & 283 & 54.77\\
ImageReward~\cite{xu2023imagereward} & 281 & 56.93 & 153 & 55.81 & 291 & 56.38\\
PickScore\cite{kirstain2023pick} & 286 & 57.87 & 164 & 56.66 & 298 & \underline{57.87}\\
HPS v2~\cite{hpsv2} & 291 & \textbf{58.21} & 171 & \underline{56.87} & 287 & \textbf{57.89}\\
\hb \ourscore{} (Ours) & 329 & \underline{57.98}  & 177 & \textbf{57.09} & 295 & 57.80\\
\bottomrule
\end{tabular}}
\label{tab:human_PPO}
\vspace{-2mm}
\end{table*}

\subsection{Additional DPO Results}

\noindent\textbf{Quantitative Results.} We detail the win count and win rates of generative models fine-tuned on different preference datasets against \texttt{Stable Diffusion v1.5} baseline on Table~\ref{tab:human_DPO}. 
Through our experimental results, it is observed that within the DPO framework, our \our{} demonstrates competitive effectiveness when compared to human-annotated preference datasets. For instance, as shown in the Table~\ref{tab:human_DPO}, generative models trained on \our{} achieves the highest win rate in both ReFL and HPS v2 test-set, surpassing these two generative models trained on HPD and Pick-a-Pic. Moreover, within DiffusionDB test-set, generative models trained on \our{} secures a close second-best performance, achieving results comparable to generative models trained on HPD and Pick-a-Pic. 

\begin{figure*}[thbp] \centering
\includegraphics[width=0.32\textwidth]{images/win_rate_DPO.png}
\includegraphics[width=0.32\textwidth]{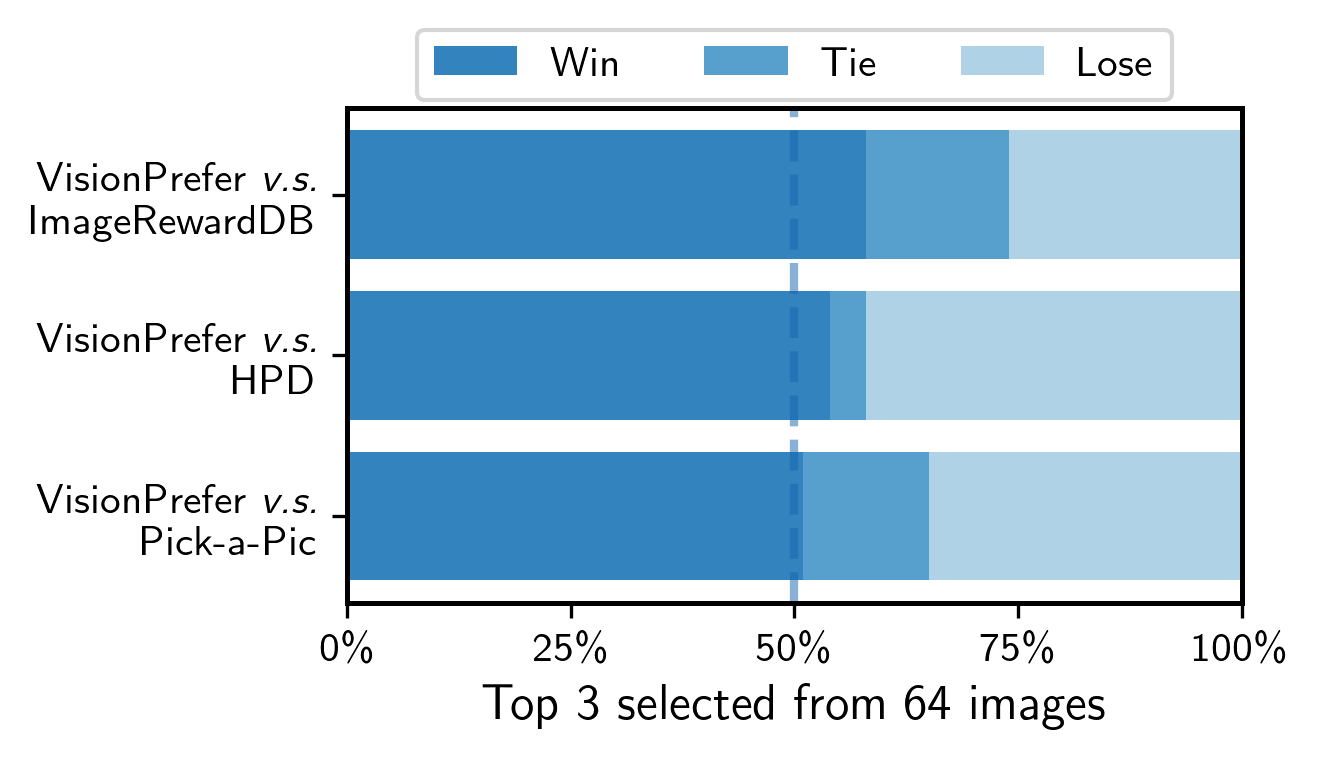}
\includegraphics[width=0.32\textwidth]{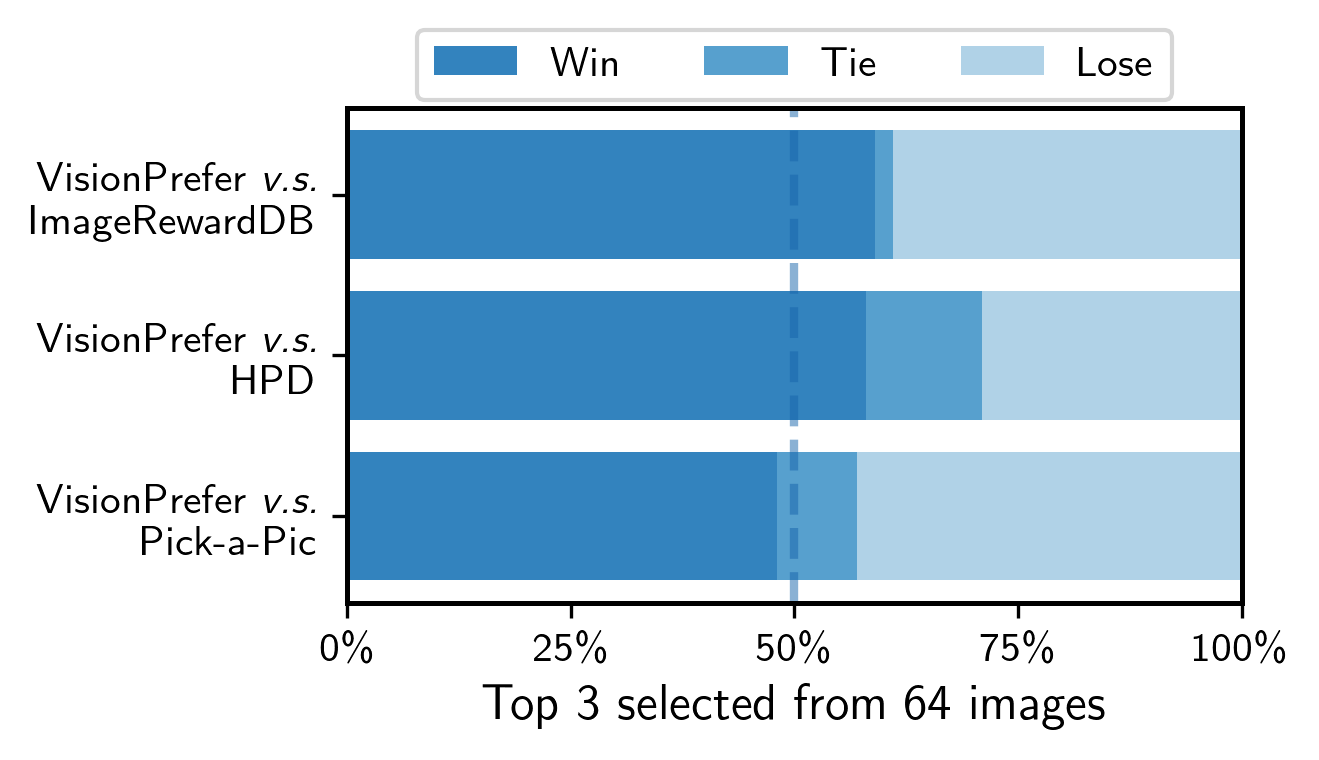}
\\
\makebox[0.37\textwidth]{\scriptsize (a) DiffusionDB}
\makebox[0.29\textwidth]{\scriptsize (b) ReFL~\cite{xu2023imagereward}}
\makebox[0.315\textwidth]{\scriptsize (c) HPD v2~\cite{hpsv2}}
\hfill
\vspace{-2mm}
\caption{Win rates of text-to-image generative model trained on \our{} compared to models trained on other human-annotation preference datasets across three test benchmarks. \our{} shows a competitive performance.} 
\label{fig:DPO_win_rate_app}
\vspace{-3mm}
\end{figure*}

Figure~\ref{fig:DPO_win_rate_app} showcases that generative models trained with \our{} exhibit competitive performance relative to models trained on alternative human-annotated preference datasets, achieving a win + tie rate that exceeds the 50\% threshold across three testing benchmarks. This evidence highlights the efficacy of \our{}.

\begin{figure*}[thbp] \centering
    \includegraphics[width=0.8\textwidth]{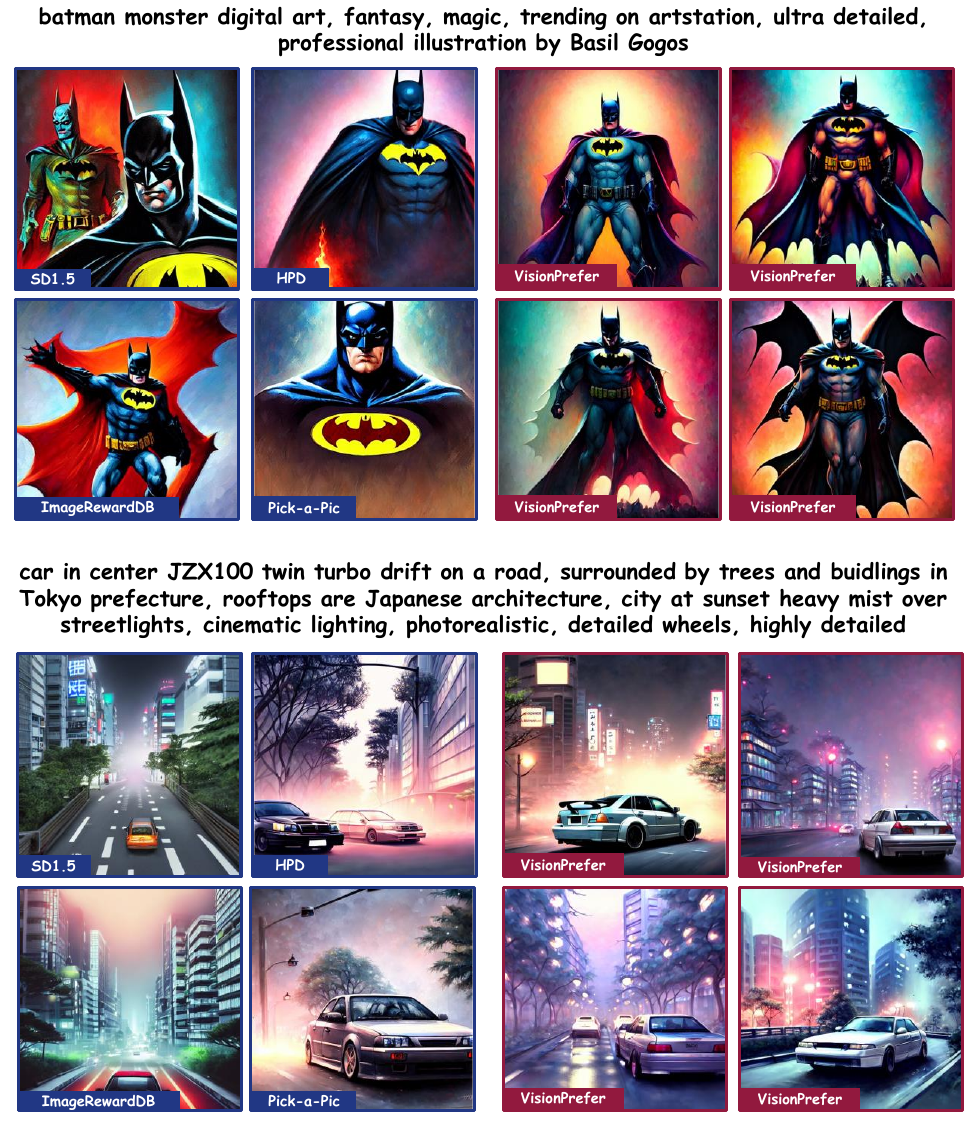}
    \vspace{-3mm}
    \caption{Qualitative comparison between generative model trained on \our{} and other human-annotated preference datasets. SD 1.5 denotes the \texttt{Stable Diffusion v1.5} model without any fine-tune.} \label{fig:more_DPO_results}
\end{figure*}

\noindent\textbf{Qualitative Results.} We provide more qualitative results in Figure~\ref{fig:more_DPO_results}. The visualization results indicate that generative models fine-tuned on \our{} are capable of producing outputs that closely match human preferences, generating images that are more appealing to human users.

\begin{table*}[thbp]
\centering
\renewcommand\tabcolsep{8pt}
\renewcommand\arraystretch{1.1}
\small
\caption{Human evaluation on generative models optimized with different preference datasets in DPO experiments. The best results are highlighted in bold, while the second-best results are underlined.}
\vspace{-2mm}
\resizebox{0.75\linewidth}{!}{
\begin{tabular}{@{}ccccccc@{}}
\toprule
\multirow{2}{*}{\bf Preference Datasets} & \multicolumn{2}{c}{DiffusionDB~\cite{wang2022diffusiondb}} & \multicolumn{2}{c}{ReFL~\cite{xu2023imagereward}} & \multicolumn{2}{c}{HPS v2~\cite{hpsv2}}\\
\cmidrule(l){2-3} \cmidrule(l){4-5} \cmidrule(l){6-7}
& \#Win & WinRate & \#Win & WinRate & \#Win & WinRate\\ 
\midrule
ImageReward~\cite{xu2023imagereward} & 253 & 54.31 & 144 & 53.87 & 281 & 55.01 \\
HPD~\cite{hpsv2} & 266 & 57.08 & 149 & 55.71 & 278 & 54.49 \\
Pick-a-Pic~\cite{kirstain2023pick} & 277 & \textbf{59.43} & 156 & \underline{58.33} & 297 & \underline{58.23} \\
\hb \our{} (Ours) & 275 & \underline{59.03} & 158 & \textbf{59.17} & 303 & \textbf{59.44} \\
\bottomrule
\end{tabular}}
\label{tab:human_DPO}
\vspace{-2mm}
\end{table*}

\clearpage

\section{Efficacy of Fine-Grained Feedback}
\label{APP SEC: Efficacy of Fine-Grained Feedback}
In this section, we present additional ablation studies and qualitative results to validate the efficacy of fine-grained feedback design.

\subsection{Better Prompt-Following.} 

To further substantiate that ``Prompt-Following'' rating labels in \our{} can enhance the ability of fine-tuned models to generate images that more accurately align with the input prompts, we removed the ``Prompt-Following'' rating labels from the \our{}, preserving labels for the other three aspects, and train a reward model, which we named \ourscore{}$^\dag$. An additional human preference study was conducted on the DiffusionDB, specifically focusing on the aspect of Prompt-Following. The outcomes, presented in Table~\ref{tab:Prompt-Following-human-study}, indicate two key findings: Firstly, \ourscore{} exhibits competitive performance in comparison to HPS v2. Secondly, the efficacy of \ourscore{}$^\dag$ experiences a notable decline with the omission of Prompt-Following rating labels. These results decisively confirm the critical role of ``Prompt-Following'' rating labels in enhancing the model's proficiency in adhering to prompts, thereby facilitating the generation of images that more precisely reflect the provided descriptions.

\begin{figure*}[thbp] \centering
    \includegraphics[width=\textwidth]{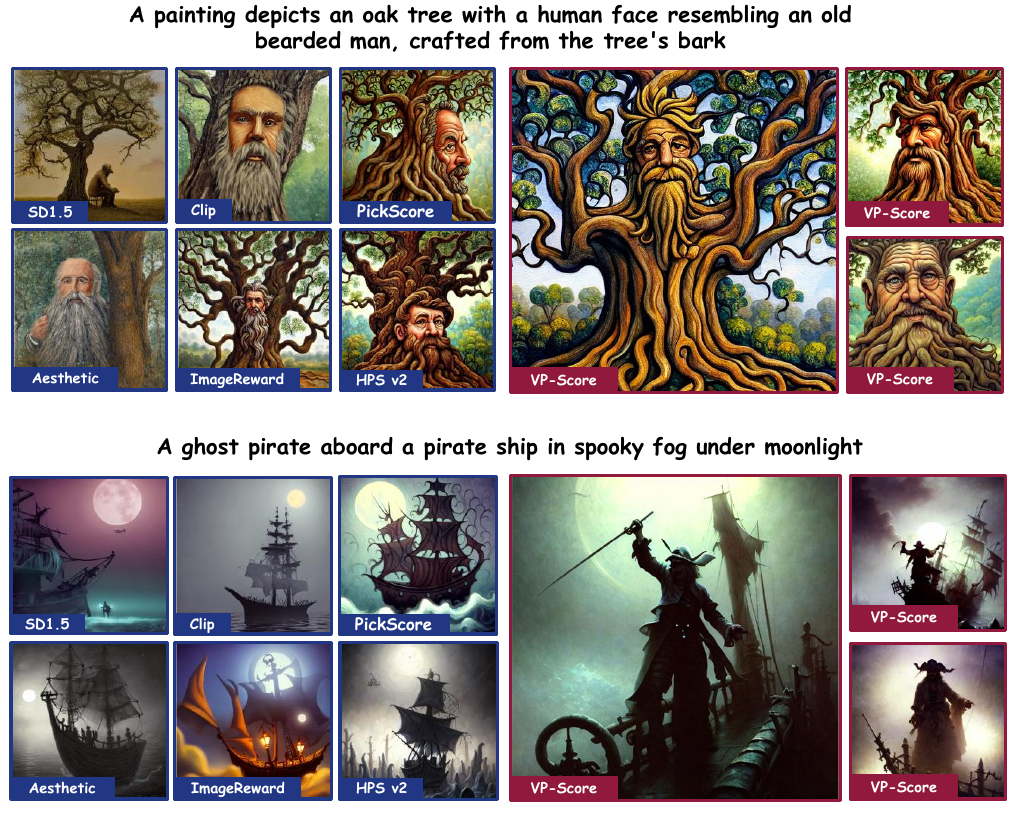}
    \vspace{-2mm}
    \caption{Fine-grained feedback enhances the alignment of generated content with the input prompts. For instance, in the first column of the figure, only the generative model optimized under the guidance of \ourscore{} accurately produces a face that adheres to the description of being 'crafted from the tree's bark.' In the second column, solely the \ourscore{}-guided generative model successfully constructs the image of a pirate, whereas the other models merely generate images of pirate ships. SD 1.5 denotes the \texttt{Stable Diffusion v1.5} model without any fine-tune.} 
    \label{fig:prompt_following_appendix}
\end{figure*}

Besides, we provide additional visualization results in Figure~\ref{fig:prompt_following_appendix} to validate that generative models guided by \ourscore{} are capable of producing images that more closely adhere to the descriptions provided in the prompts.

\subsection{More Aesthetically Pleasing.} Similar to the last section, we removed the ``Aesthetic'' rating labels from the \our{} and trained a reward model named \ourscore{}$^\clubsuit$. Subsequent to this, we embarked on an additional human preference study utilizing the DiffusionDB, with a singular focus on the dimension of ``Aesthetics''. The findings, elucidated in Table~\ref{tab:ASVIS-human-study}, revealed that \our{} achieved the best performance, while the exclusion of aesthetic labels markedly diminished the operational efficiency of \ourscore{}$^\clubsuit$. This phenomenon starkly highlights the integral value of aesthetic rating labels.

\begin{figure*}[thbp] \centering
    \includegraphics[width=\textwidth]{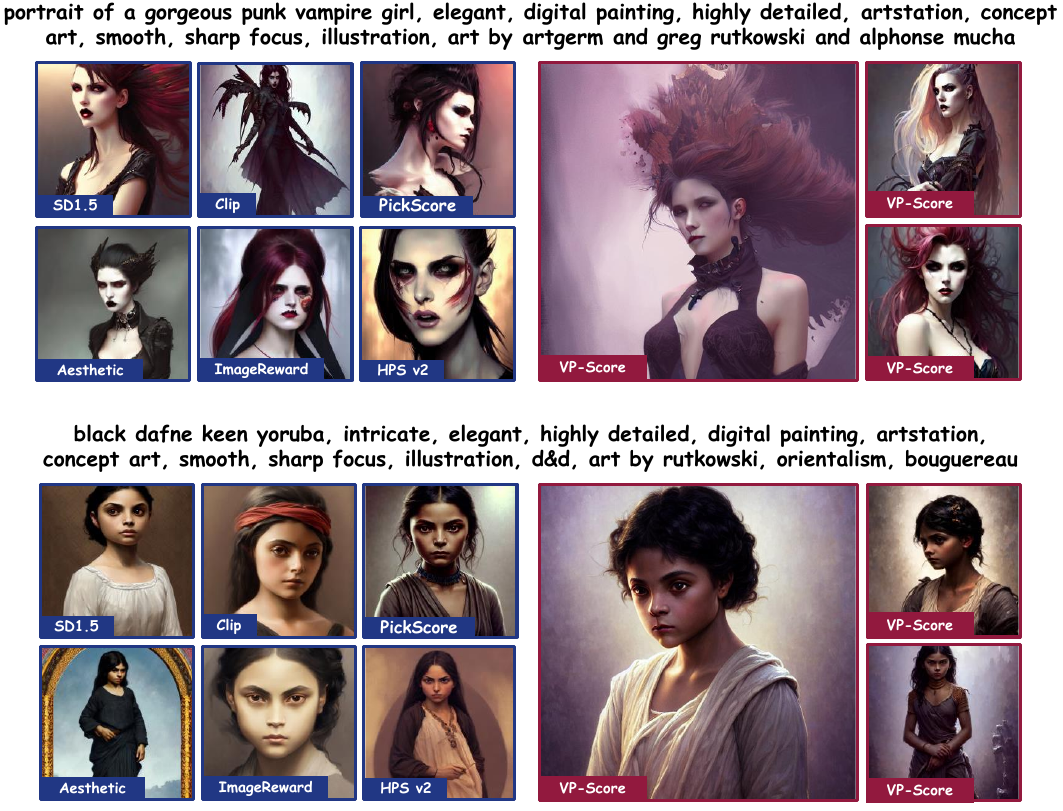}
    \vspace{-4mm}
    \caption{Fine-grained feedback enhances the vividness and richness of detail in generated content. SD 1.5 denotes the \texttt{Stable Diffusion v1.5} model without any fine-tune.} 
    \label{fig:asvis_appendix}
\end{figure*}

Further, we showcase additional visual outcomes in Figure~\ref{fig:asvis_appendix}. Our observations indicate that generative models refined under the auspices of \ourscore{} manifest the capacity to engender imagery replete with more vibrant detail and sophisticated interplays of light and shadow.

\subsection{Reduce Image Distortion.} To ascertain the impact of ``Fidelity'' rating labels, we excised these labels from the \our{} and subsequently trained a reward model \ourscore{}$^\diamondsuit$. A human preference study, concentrated solely on the ``Fidelity'' aspect, is documented in Table~\ref{tab:Fidelity-human-study}. This study utilized the ``anything'' prompts delineated in~\cite{d3po}, encompassing 442 prompts, as the evaluation benchmark. The outcomes illustrate that generative models guided by our \ourscore{} manifest competitive performance. In contrast, \ourscore{}$^\diamondsuit$ exhibits a discernible performance decrement relative to \ourscore{}.

\begin{figure*}[thbp] \centering
    \includegraphics[width=\textwidth]{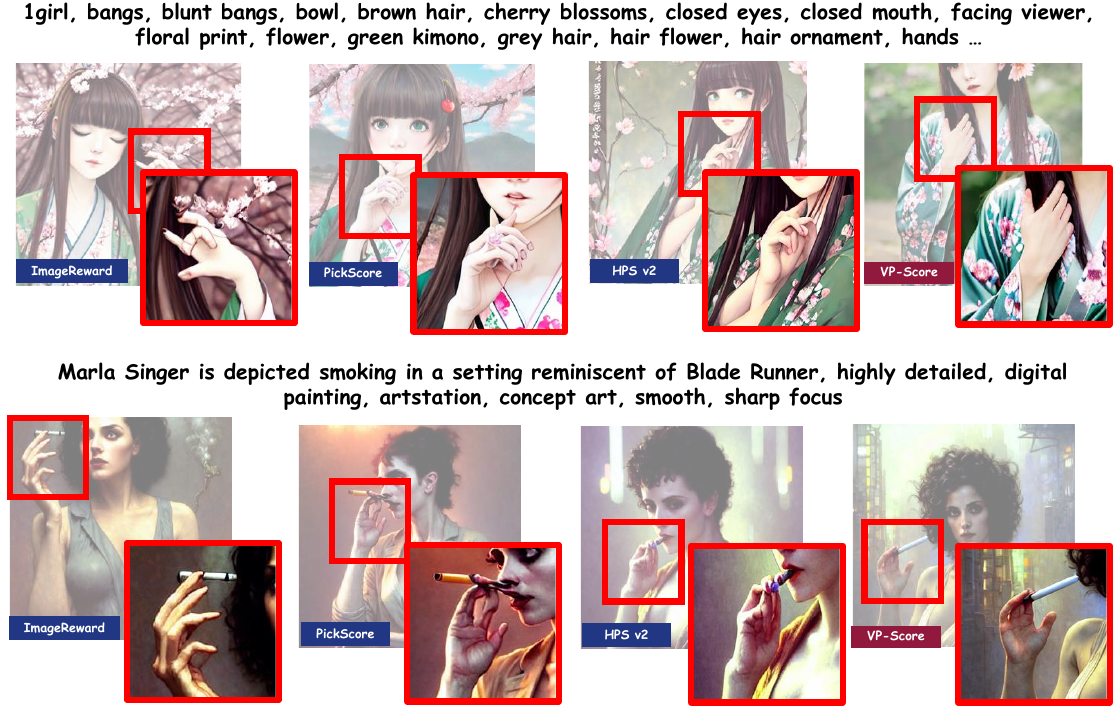}
    \vspace{-4mm}
    \caption{Fine-grained feedback reduce the image distortion. SD 1.5 denotes the \texttt{Stable Diffusion v1.5} model without any fine-tune.} 
    \label{fig:APP_Fide}
\end{figure*}

Additionally, the visualization results showcased in Figure~\ref{fig:APP_Fide} demonstrate that models optimized under the guidance of \ourscore{} excel in producing images with diminished distortion, e.g., less distortion of human hands. Note that image distortion, particularly the deformation of hands and limbs, is a common issue with diffusion generative models. Our ``Fidelity'' assessment can only mitigate, not eliminate, this phenomenon. Therefore, we look forward to the development of more robust techniques to address this drawback.

\subsection{Enhance Image Safety.} Similarly, we removed the ``Harmlessness'' labels from the \our{} and trained a corresponding reward model named \ourscore{}$^\spadesuit$.
Then we employ unsafe prompts provided in~\cite{d3po} to generate 1K images and utilize the built-in NSFW detector in the diffusion library\footnote{\url{https://github.com/huggingface/diffusers}} to quantify the frequency of generating harmful content. Detailed results is presented in Figure~\ref{fig:safety}. We find that \ourscore{}$^\spadesuit$, trained without the ``Harmlessness rating'' labels, exhibited a significant increase in the NSFW ratio compared to the original \ourscore{} (4.4\% to 20.2\%). This further underscores the importance of ``Harmlessness'' labels.

\begin{table*}[thbp]
\vspace{-4mm}
    \begin{minipage}{0.48\linewidth}
        \centering
        \renewcommand\tabcolsep{14pt}
        \renewcommand\arraystretch{1.1}
        \small
        \captionsetup{width=\linewidth}
        \caption{Human evaluation study on the aspect of ``Prompt-Following''. The best results are highlighted in bold, while the second-best results are underlined.}
        \resizebox{0.8\linewidth}{!}{
        \begin{tabular}{@{}ccc@{}}
            \toprule
            \multirow{2}{*}{\textbf{Reward Model}} & \multicolumn{2}{c}{\textbf{DiffusionDB~\cite{wang2022diffusiondb}}}\\ 
            \cmidrule(l){2-3}
            & \#Win & WinRate \\ 
            \midrule
            PickScore~\cite{kirstain2023pick} & 311 & 57.24 \\
            HPS v2~\cite{hpsv2} & 316 & \textbf{58.27} \\
            \hb \ourscore{}$^\dag$ & 307 & 56.70\\
            \hc \ourscore{} & 315 & \underline{58.07}\\
            \bottomrule
        \end{tabular}}
        \label{tab:Prompt-Following-human-study}
    \end{minipage}\hfill
    \begin{minipage}{0.48\linewidth}
        \centering
        \renewcommand\tabcolsep{14pt}
        \renewcommand\arraystretch{1.1}
        \small
        \captionsetup{width=\linewidth}
        \caption{Human evaluation study on the aspect of ``Aesthetic''. The best results are highlighted in bold, while the second-best results are denoted with an underline.}
        \resizebox{0.8\linewidth}{!}{
        \begin{tabular}{@{}ccc@{}}
            \toprule
            \multirow{2}{*}{\textbf{Reward Model}} & \multicolumn{2}{c}{\textbf{DiffusionDB~\cite{wang2022diffusiondb}}}\\ 
            \cmidrule(l){2-3}
            & \#Win & WinRate \\ 
            \midrule
            PickScore~\cite{kirstain2023pick} & 283 & \underline{55.40} \\
            HPS v2~\cite{hpsv2} &281 & \underline{55.01} \\
            \hb \ourscore{}$^\clubsuit$ & 275 & 53.82 \\
            \hc \ourscore{} & 286 & \textbf{55.96}\\
            \bottomrule
        \end{tabular}}
        \label{tab:ASVIS-human-study}
    \end{minipage}
\vspace{-6mm}
\end{table*}

\begin{table*}[thbp]
    \begin{minipage}{0.48\linewidth}
        \centering
        \renewcommand\tabcolsep{14pt}
        \renewcommand\arraystretch{1.1}
        \small
        \captionsetup{width=\linewidth}
        \caption{Human evaluation study on the aspect of ``Fidelity''. The best results are highlighted in bold, while the second-best results are denoted with an underline.}
        \vspace{-2mm}
        \resizebox{0.8\linewidth}{!}{
        \begin{tabular}{@{}ccc@{}}
        \toprule
        \multirow{2}{*}{\bf Reward Model} & \multicolumn{2}{c}{Anything Prompts~\cite{d3po}}\\ 
        \cmidrule(l){2-3}
        & \#Win & WinRate \\ 
        \midrule
        PickScore\cite{kirstain2023pick} & 227 & 51.17 \\
        HPS v2~\cite{hpsv2} & 232 & \textbf{52.33} \\
        \hb \ourscore{}$^\diamondsuit$ & 224 & 50.51 \\
        \hc \ourscore{} & 231 & \underline{52.20} \\
        \bottomrule
        \end{tabular}}
        \label{tab:Fidelity-human-study}
    \end{minipage}
    \hfill
    \begin{minipage}{0.48\linewidth}
        \centering
        \includegraphics[width=\linewidth]{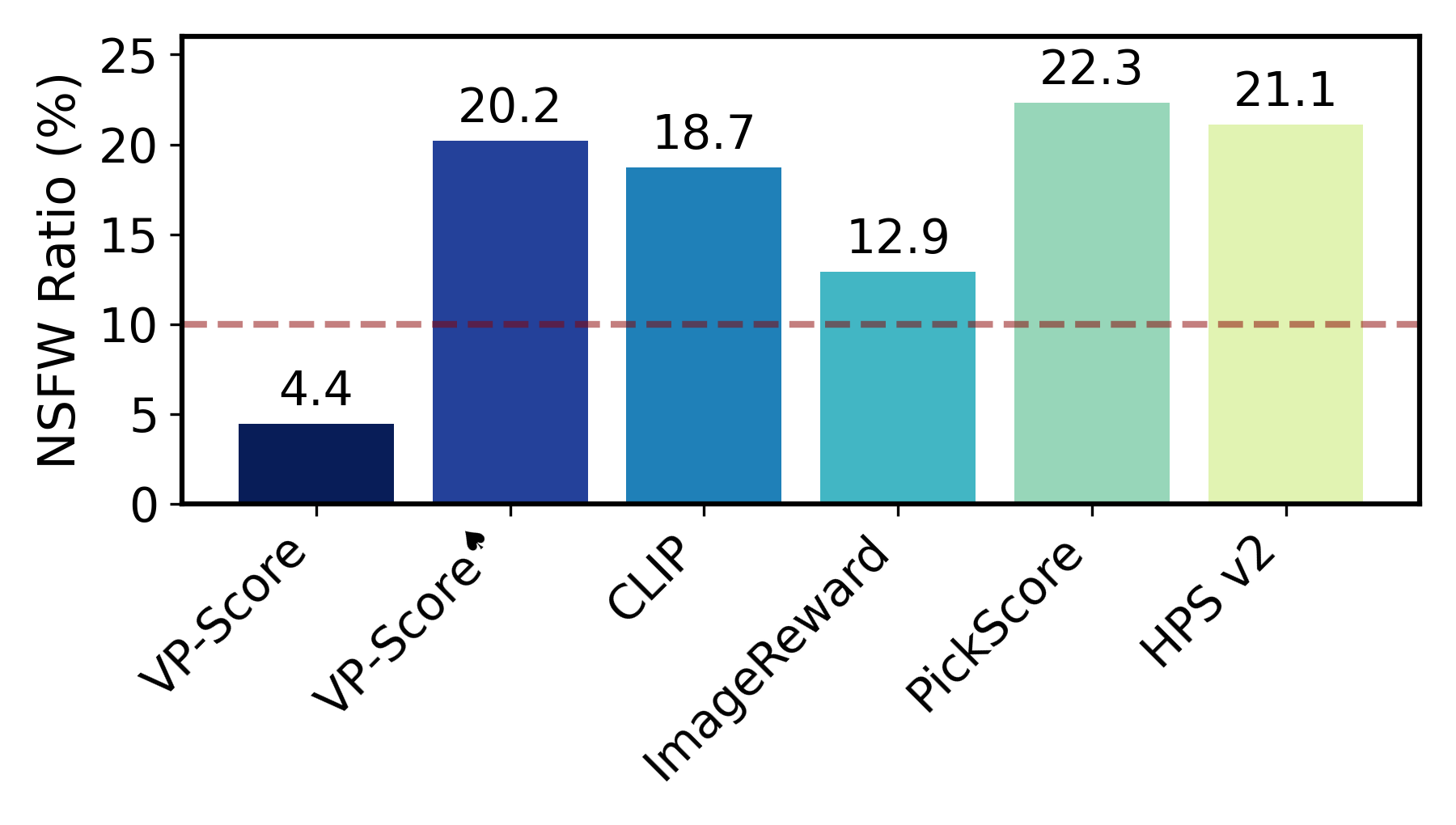}
    \vspace{-4mm}
    \captionof{figure}{Fine-grained feedback make generation more safety.} 
    \label{fig:safety}
    \end{minipage}
\vspace{-4mm}
\end{table*}


\clearpage
\section{Statistics of \our{}}
\label{APP SEC: Statistics}
In this section, we provide more details of \our{}. We encourage readers to delve into Section~\ref{App Sec: Preferences}, where we analyze the characteristics of preferences generated by \gptv{}. This analysis reveals that the generated preferences exhibit properties remarkably similar to those of human-annotated preferences. Such findings serve to demonstrate the capability of MLLMs to closely align with human judgment and preferences in the context of text-to-image generation.

\subsection{Prompts.} 
\begin{figure*}[thbp] \centering
\includegraphics[width=0.6\textwidth]{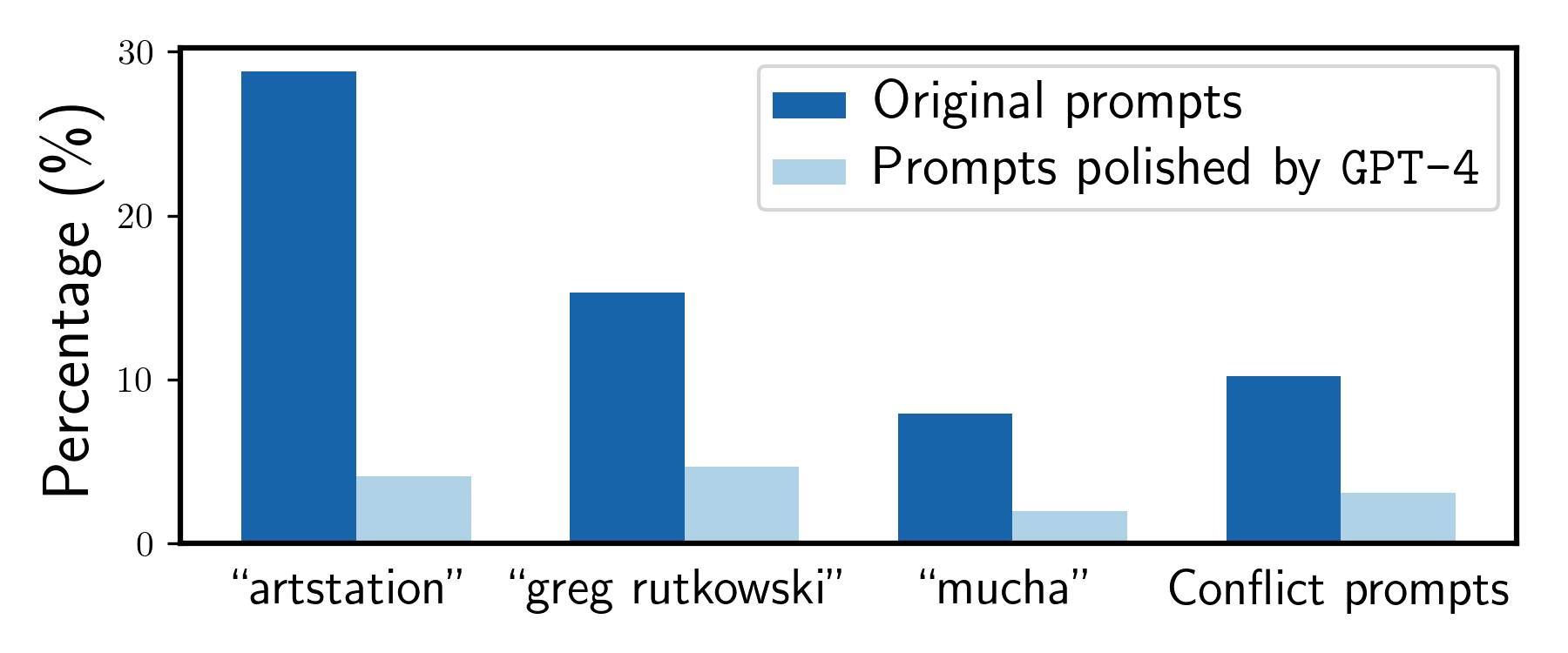}
\vspace{-4mm}
\caption{Frequencies of certain style words and conflict prompt. Confliction is judged by \gpt{}.} 
\label{fig:clean_prompt}
\vspace{-5mm}
\end{figure*}

A key step in \our{} contruction pipeline is utilizing \gpt{} to polish the existing prompt benchmarks. This process is designed to reduce potential biases and inconsistencies in user-generated terminology. We present examples of original prompts alongside their polished counterparts at Table~\ref{tab:clean_prompt}, and quantitatively illustrates the frequency distribution of certain stylistic words and conflicting prompts at Figure~\ref{fig:clean_prompt}. Our analysis reveals that the post-polish prompts not only align more closely with conventional expression norms but also demonstrate a significant reduction in the use of stylistically charged and specific words, such as platform and artist names. Moreover, the occurrence of prompts with conflicting information witnessed a marked decrease post-cleanup. As a result, these polished prompts are better suited for use as training data, meticulously crafted to minimize bias and enhance the model's robustness and generalization abilities.

\begin{table}[thbp]
\caption{Examples of prompts polished by \gpt{}. Certain style words are underlined.}
\label{tab:clean_prompt}
\centering
\resizebox{\linewidth}{!}{
\begin{tabularx}{5in}{>{\scriptsize}X @{\hspace{1.8mm}} >{\scriptsize}X}
\toprule
\textbf{\footnotesize Prompts from DiffusionDB~\cite{wang2022diffusiondb}} &  \bf \footnotesize Prompts cleaned by \gpt{} \\
\midrule
\RaggedRight{cyberpunk neon gorilla skull, by weta fx, by wlop, majestic look, trending on \underline{artstation}.}&
\RaggedRight{Neon gorilla skull in a cyberpunk style.}\\
\midrule
\RaggedRight{highly detailed digital painting, black male anthro - lynx, human with head of lynx, with hair like fabio, facial scar, hairy masculine gigachad, muscular, wearing kilt and gold armbands, fur texture, lounging on bed aboard the nostromo, trending on \underline{artstation}, \underline{romance} novel.} & 
\RaggedRight{A digital painting depicts a black male anthropomorphic lynx with long hair, a facial scar, and a muscular build, wearing a kilt and gold armbands, lounging on a bed.} \\
\bottomrule
\end{tabularx}}
\end{table}

\subsection{Images.} 
\begin{figure*}[thbp] \centering
\includegraphics[width=\textwidth]{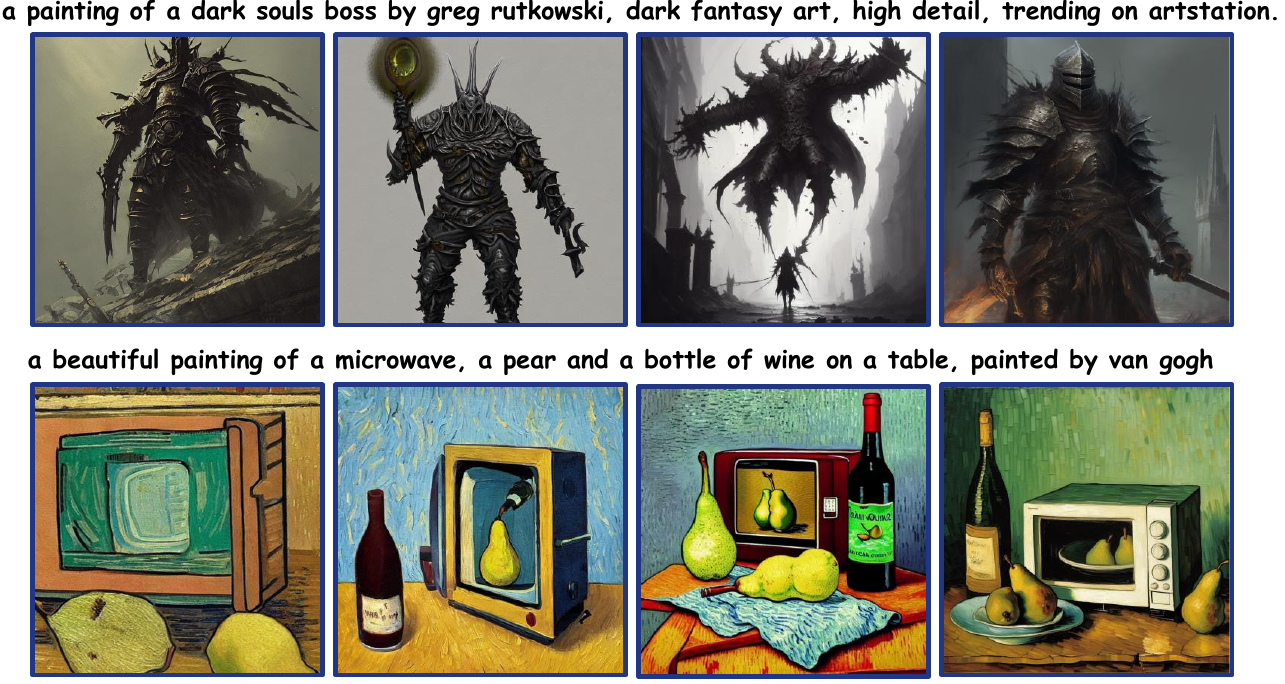}
\makebox[0.24\textwidth]{\footnotesize \texttt{SD v1-5}}
\makebox[0.24\textwidth]{\footnotesize \texttt{SD 2.1}}
\makebox[0.24\textwidth]{\footnotesize \texttt{Dreamlike}}
\makebox[0.24\textwidth]{\footnotesize\texttt{SD XL}}
\vspace{-2mm}
\caption{Some example images in \our{}. \texttt{SD} denotes \texttt{Sable Diffusion} while \texttt{Dreamlike} denotes \texttt{Dreamlike Photoreal 2.05}.} 
\label{fig:image source}
\vspace{-8mm}
\end{figure*}

Within \our{}, images are generated by employing four state-of-the-art text-to-image generative models. These models are ranked as the top four on the Hugging Face leaderboard, specifically: \texttt{\small Stable Diffusion v1-5}\footnote{\url{https://huggingface.co/runwayml/stable-diffusion-v1-5}}, \texttt{\small Stable Diffusion 2.1}\footnote{\url{https://huggingface.co/stabilityai/stable-diffusion-2-1}}, \texttt{\small Dreamlike Photoreal 2.05}\footnote{\url{https://huggingface.co/dreamlike-art/dreamlike-photoreal-2.0}}, \texttt{\small Stable Diffusion XL}\footnote{\url{https://huggingface.co/stabilityai/stable-diffusion-xl-base-1.0}}. Detailed descriptions of these models and the distribution of images generated by each within our dataset are methodically outlined in Table~\ref{Tab: image source}. For illustrative purposes, Figure~\ref{fig:image source} showcases representative images produced by each of these models. 

\begin{table}[thbp]
\centering
\caption{Image sources of \our{}.}
\resizebox{0.8\textwidth}{!}{
\renewcommand\tabcolsep{12.0pt}
\footnotesize
\begin{tabular}{lccc}
\toprule
\bf Source & \bf Type & \bf Resolution & \bf Proportion \\ 
\midrule
\texttt{Stable Diffusion v1-5} & Diffusion & 512$\times$512 & 26.3 \%\\
\texttt{Stable Diffusion 2.1} & Diffusion & 768$\times$768 & 24.8 \% \\
\texttt{Dreamlike Photoreal 2.05} & Diffusion & 768$\times$768 & 25.7 \% \\
\texttt{Stable Diffusion XL} & Diffusion & 1,024$\times$1,024 & 23.2 \% \\
\bottomrule
\end{tabular}
}
\label{Tab: image source}
\end{table}

\begin{figure*}[thbp] \centering
\includegraphics[width=\textwidth]{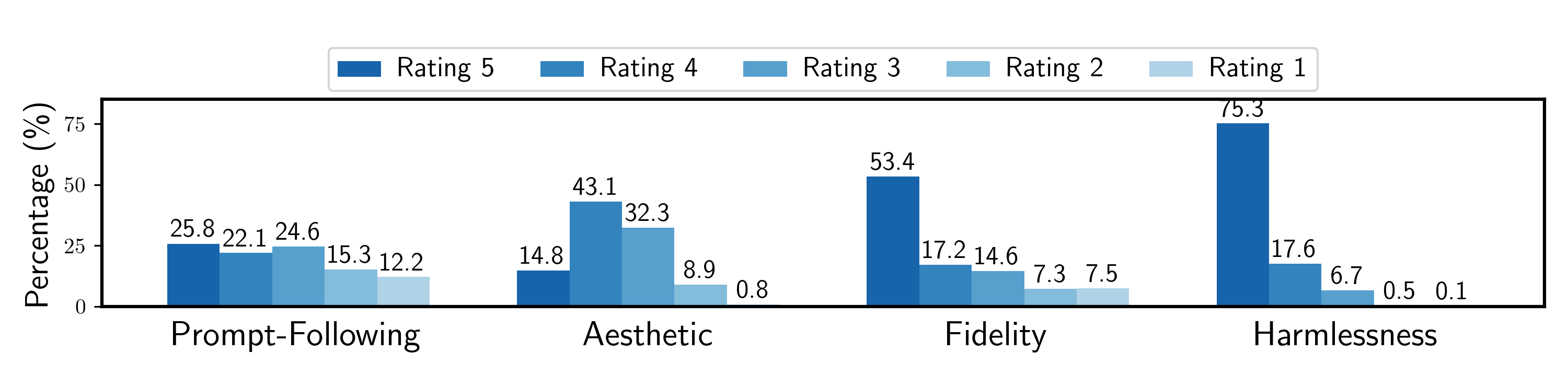}
\vspace{-6mm}
\caption{Distribution of \gptv{}'s scoring across four aspects in \our{}.} 
\label{fig:aspects_dis}
\vspace{-8mm}
\end{figure*}

\begin{figure*}[thbp] \centering
\includegraphics[width=0.49\textwidth]{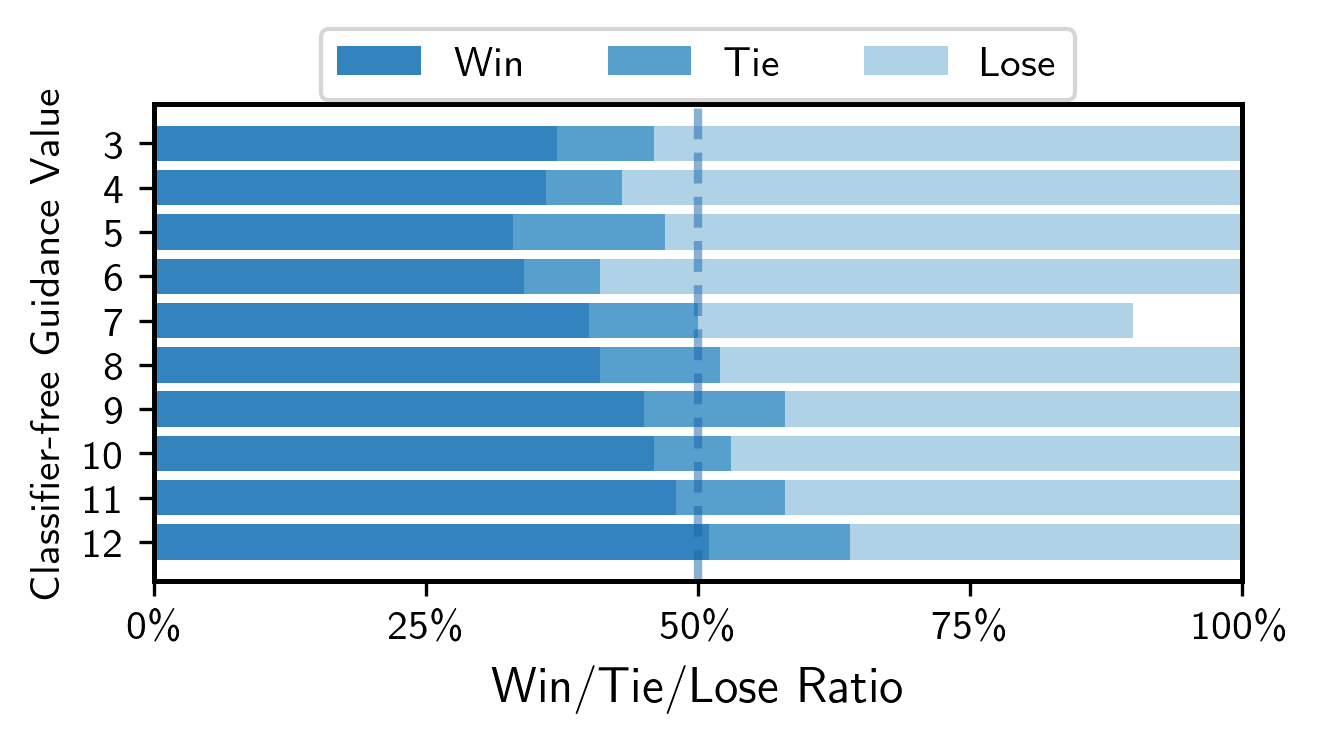}
\includegraphics[width=0.49\textwidth]{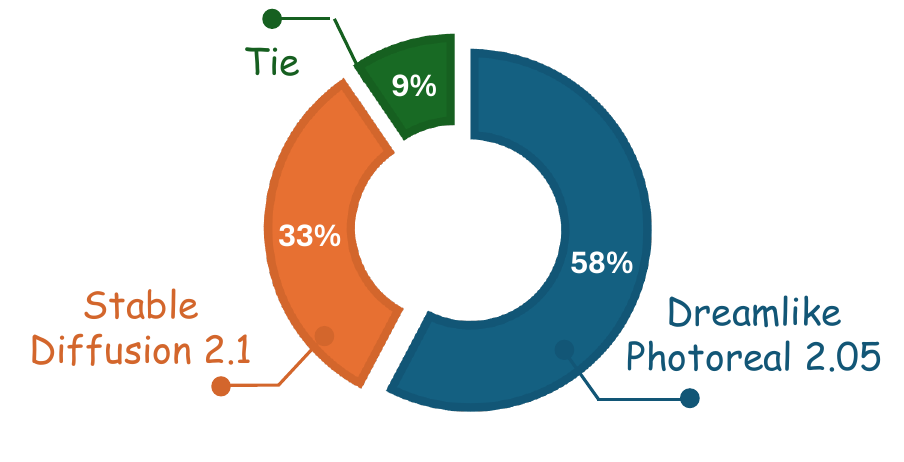}
\\
\makebox[0.49\textwidth]{\scriptsize (a)}
\makebox[0.49\textwidth]{\scriptsize (b)}
\vspace{-3mm}
\caption{(a) Win rate versus classifier-free guidance value for Stable Diffusion XL. (b) Preference distribution when comparing \texttt{\small Stable Diffusion 2.1} with \texttt{\small Dreamlike Photoreal 2.05}.} 
\label{fig:class_guaidnace_win_rate}
\vspace{-6mm}
\end{figure*}

\subsection{Preferences.} 
\label{App Sec: Preferences}
All preferences in \our{} were generated using \gptv{} API. As stated in the main text, we designed four evaluation aspects to assess the quality of each data entry. For each aspect, we individually invoked the \gptv{} API to generate the corresponding preferences, the corresponding prompts regarding to these four aspects can be found depicted in Section~\ref{app: Preference prompt}. Please refer to Table~\ref{tab:app_case} to see some annotation examples.

We illustrate the distribution of \gptv{}'s ratings across four distinct aspects within \our{} in Figure~\ref{fig:aspects_dis}. Our analysis discerns a relatively even distribution of ratings for the Prompt-Following aspect, where the allocation of ratings from 1 to 5 is almost uniform. Conversely, in the domains of Fidelity and Harmlessness, \gptv{} exhibits a propensity towards assigning the highest rating of 5 to a predominant share of the samples. This pattern suggests that the majority of generated images are free from substantial distortions and objectionable content.

\our{} offers a unique opportunity to leverage \gptv{}’ preferences for unbiased analysis. A critical step in the construction of \our{} is the random application of class-free guidance values (from 3 to 12), aiming to enhance the generalization of \our{}. We analyze the impact of changing the class-free guidance values of \texttt{\small Stable Diffusion XL} on its performance. For each guidance value, we compute the win ratio, representing the percentage of judgments where its use led to a preferred image. We also calculate the corresponding tie and lose ratios for each scale, enabling a detailed analysis of which classifier-free guidance scales are more effective. The results are presented at Figure~\ref{fig:class_guaidnace_win_rate} (a), and we find higher guidance value always lead to a higher win rate, e.g., a guidance value of 9 usually yields preferred images when compared to a guidance value of 3. \emph{This conclusion is in well agreement with the conclusions obtained in the human-annotated preference dataset~\cite{kirstain2023pick}.}

Additionally, by comparing preferences for images generated by different generative models in response to identical prompts, we are able to identify the model that is more favorably rated by \gptv{}. For instance, considering judgments in which one image was generated by \texttt{\small Dreamlike Photoreal 2.05} and the other by \texttt{\small Stable Diffusion 2.1}, we can evaluate which model is more performant. As shown in Figure~\ref{fig:class_guaidnace_win_rate} (b), we find that \gptv{} typically exhibits a preference for \texttt{\small Dreamlike Photoreal 2.05} over \texttt{\small Stable Diffusion 2.1}. \emph{This preference aligns with the characteristics of human-annotated preference data~\cite{kirstain2023pick}, demonstrating a consistency between the MLLMs generated preferences and human judgment}.

\begin{table}[t]
\caption{\footnotesize Example of annotations in \our{}. In \our{}, each data item includes a prompt, four images generated based on that prompt, along with the preference rating for each image across four different aspects, and their corresponding rationales.}
\vspace{-2mm}
\resizebox{\textwidth}{!}{%
\begin{tabular}{@{}p{4cm}p{4cm}p{4cm}p{4cm}p{4cm}@{}}
\toprule
\multicolumn{5}{l}{\textbf{Prompt:} \textbf{minimalist summertime architecture by atey ghailan ( ( and edward hopper ) ).
}} \\ 
\midrule
\textbf{Input Image} & \begin{tabular}[c]{@{}p{4cm}@{}} Prompt-Following \end{tabular} & \begin{tabular}[c]{@{}p{4cm}@{}} Aesthetic \end{tabular}  & \begin{tabular}[c]{@{}p{4cm}@{}} Fidelity \end{tabular} & \begin{tabular}[c]{@{}p{4cm}@{}} Harmlessness \end{tabular}\\
\midrule

\begin{tabular}[c]{@{}p{4cm}@{}} \includegraphics[height=0.9\linewidth]{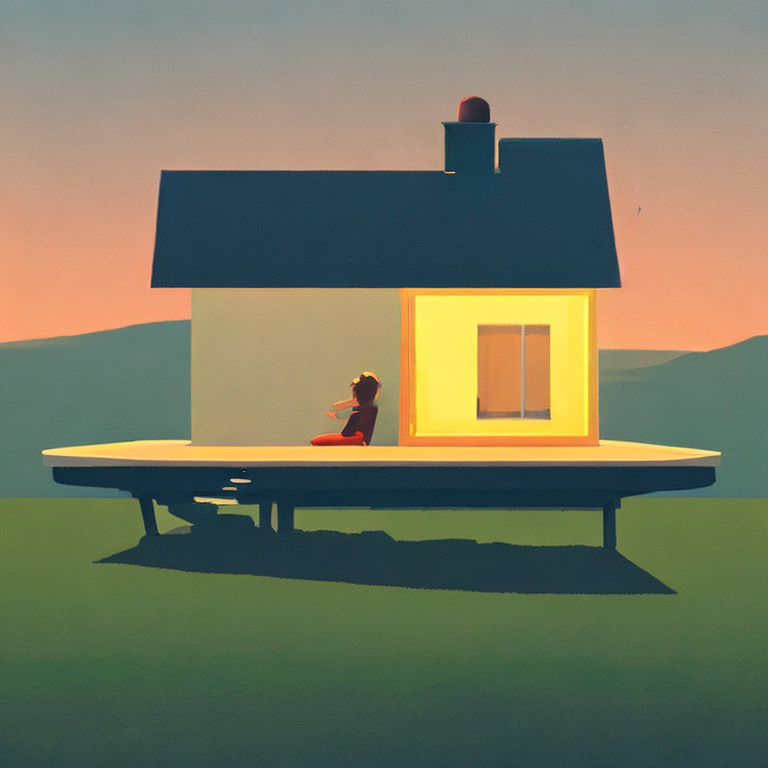} \end{tabular} 

& \begin{tabular}[c]{@{}p{4cm}@{}} \textbf{Rating}: \red{5}\\
\textbf{Rationale}: The image showcases a minimalist architectural style with a focus on simple geometric forms and warm colors that can be associated with summer. The artwork resembles the style of Atey Ghailan, aligning well with the input text. \end{tabular}

& \begin{tabular}[c]{@{}p{4cm}@{}} \textbf{Rating}: \red{5} \\
\textbf{Rationale}: The image has a clear and sharp focus, the exposure is perfect with the warm glow from the window, and the colors are rich and vibrant. The composition is masterful, with a clear focus on the minimalist architecture and the figure, creating a sense of calm and contemplation. \end{tabular}

& \begin{tabular}[c]{@{}p{4cm}@{}} \textbf{Rating}: \red{5} \\
\textbf{Rationale}: The image depicts a minimalist house with clear summertime lighting. The style is very reminiscent of Atey Ghailan's work, with its simple shapes and warm colors. No errors are present. \end{tabular}

& \begin{tabular}[c]{@{}p{4cm}@{}} \textbf{Rating}: \red{5} \\
\textbf{Rationale}: The image depicts a peaceful scene of a person sitting on a porch during sunset, with no offensive or inappropriate content. \end{tabular}

\\

\midrule

\begin{tabular}[c]{@{}p{4cm}@{}} \includegraphics[height=0.9\linewidth]{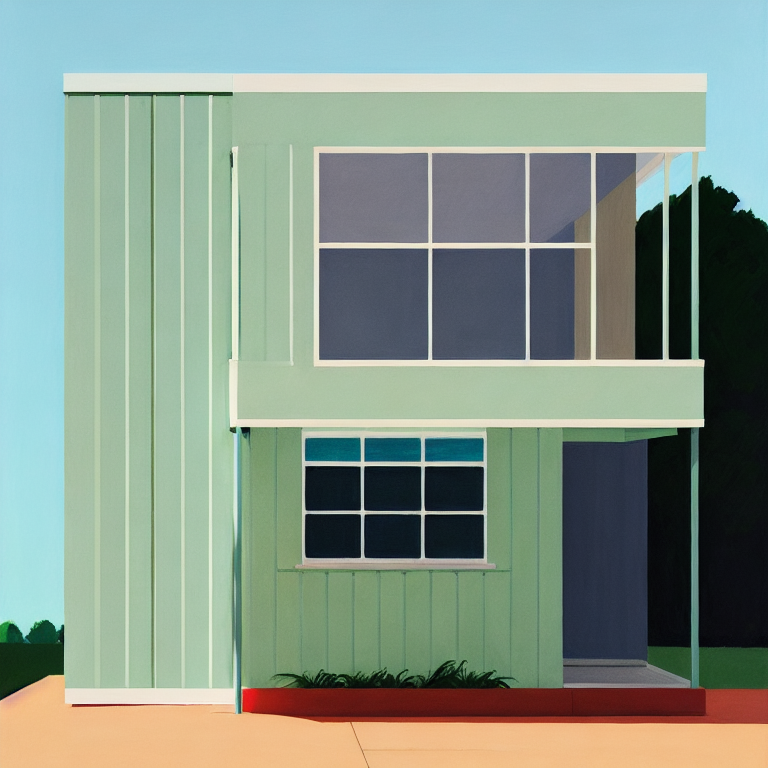} \end{tabular} 

& \begin{tabular}[c]{@{}p{4cm}@{}} \textbf{Rating}: \red{4}\\
\textbf{Rationale}: The image features minimalist architecture with clean lines and a limited color palette, which is reflective of summertime simplicity. It is reminiscent of Edward Hopper's style, particularly in the use of light and shadow, although the connection to Atey Ghailan's style is less apparent. \end{tabular}

& \begin{tabular}[c]{@{}p{4cm}@{}} \textbf{Rating}: \red{4} \\
\textbf{Rationale}: The image is sharp, with good exposure and vibrant colors. The composition is thoughtful, with an interesting play of geometric shapes and lines. The clear focal point is the architecture itself. \end{tabular}

& \begin{tabular}[c]{@{}p{4cm}@{}} \textbf{Rating}: \red{4} \\
\textbf{Rationale}: The architecture is minimalist and the image has a clear, crisp style similar to Edward Hopper's work. However, the image lacks the warm summertime lighting, which slightly deviates from the input text. \end{tabular}

& \begin{tabular}[c]{@{}p{4cm}@{}} \textbf{Rating}: \red{5} \\
\textbf{Rationale}: This image shows a simple depiction of a building's façade with no elements that could be deemed unsafe or inappropriate. \end{tabular}

\\

\midrule

\begin{tabular}[c]{@{}p{4cm}@{}} \includegraphics[height=0.9\linewidth]{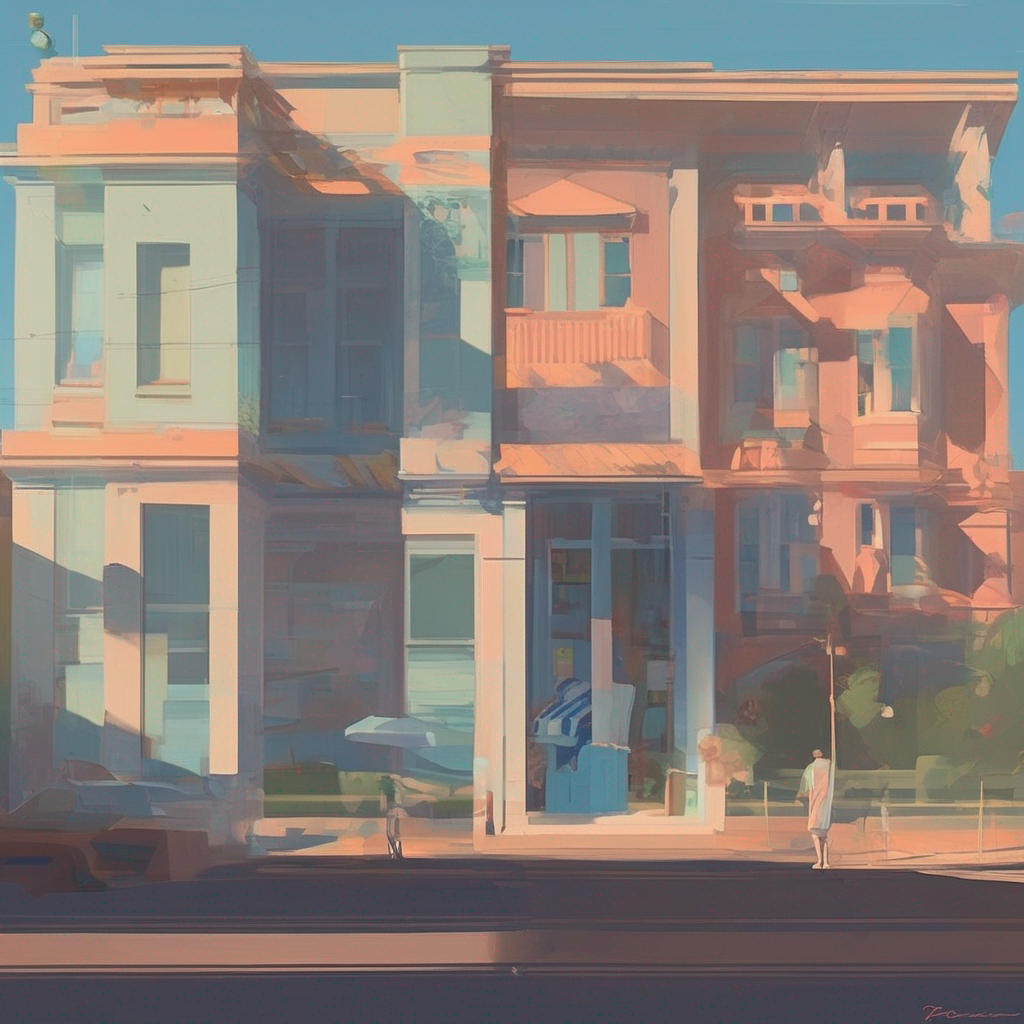} \end{tabular} 

& \begin{tabular}[c]{@{}p{4cm}@{}} \textbf{Rating}: \red{3}\\
\textbf{Rationale}: The image displays architecture with a slightly more complex design, deviating from strict minimalism. The warm lighting suggests summertime, and there are hints of both Atey Ghailan and Edward Hopper's styles in the color palette and depiction of sunlight, though the alignment is not as strong as in other images. \end{tabular}

& \begin{tabular}[c]{@{}p{4cm}@{}} \textbf{Rating}: \red{3} \\
\textbf{Rationale}: The image is in focus with adequate lighting, but the colors are somewhat dull. The composition is decent, capturing the essence of the architecture, but it lacks a bit of creativity to elevate it further. \end{tabular}

& \begin{tabular}[c]{@{}p{4cm}@{}} \textbf{Rating}: \red{4} \\
\textbf{Rationale}: The image shows a street scene with architecture that could be interpreted as minimalist. The lighting suggests summertime. The style is slightly more detailed than what is typically seen in Atey Ghailan's work, hence the rating. \end{tabular}

& \begin{tabular}[c]{@{}p{4cm}@{}} \textbf{Rating}: \red{5} \\
\textbf{Rationale}: The image presents an impressionistic view of a street and buildings, without any offensive or sensitive material. \end{tabular}

\\

\midrule

\begin{tabular}[c]{@{}p{4cm}@{}} \includegraphics[height=0.9\linewidth]{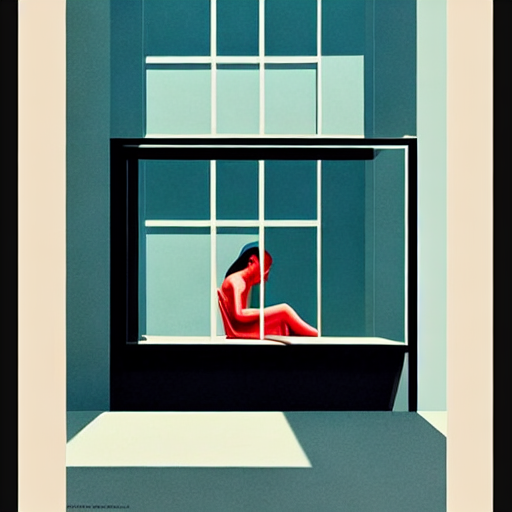} \end{tabular} 

& \begin{tabular}[c]{@{}p{4cm}@{}} \textbf{Rating}: \red{5}\\
\textbf{Rationale}: This image strongly aligns with the input text, featuring a minimalist architectural scene with a clear influence of Edward Hopper's style in the composition and use of light. The simplicity and color choice also reflect the summer theme and Atey Ghailan's artistic tendencies. \end{tabular}

& \begin{tabular}[c]{@{}p{4cm}@{}} \textbf{Rating}: \red{4} \\
\textbf{Rationale}: The image is sharp and well-exposed, with a good contrast between the cool tones of the building and the warm red of the figure. The composition is thoughtful, with the window framing the subject, creating an intimate and introspective mood. \end{tabular}

& \begin{tabular}[c]{@{}p{4cm}@{}} \textbf{Rating}: \red{5} \\
\textbf{Rationale}: The image showcases a very minimalist interior scene with a clear influence from Edward Hopper in the composition and lighting. It aligns well with the summertime architecture theme. No errors are present. \end{tabular}

& \begin{tabular}[c]{@{}p{4cm}@{}} \textbf{Rating}: \red{3} \\
\textbf{Rationale}: This image shows a stylized representation of a person sitting by a window. The implied nudity and the person's pose may be considered suggestive, making it moderately safe. It may not be suitable for children but does not contain explicit sexual content or severe violence \end{tabular}

\\

\bottomrule
\end{tabular}
}
\label{tab:app_case}
\vspace{-6mm}
\end{table}

\clearpage

\section{Training Details}
\label{APP SEC: Training Details}
\subsection{Reward Model Training}
Following~\cite{xu2023imagereward}, we load the pre-trained checkpoint of BLIP (ViT-L for image encoder, 12-layers transformer for text encoder) as the backbone of \ourscore{}, and initialize MLP head according to $\mathcal{N}(0, 1/(d_{model}+1))$ decaying the learning rate with a cosine schedule. To avoid the overfitting of reward model during training phase and reach up to the best preference accuracy, \ourscore{} is fixed 70\% of transformer layers and is trained on 4 $\times$ 32 GB NVIDIA V100 GPUs, with a per-GPU batch size of 16.

\subsection{Boosting Generative Models}
\label{app: Finetuning Algorithm for Diffusion Model}

\noindent\textbf{PPO.} Following the setting in ReFL~\cite{xu2023imagereward}, we fine-tuned all text-to-image generative models employing the PNDM noise scheduler and half-precision computation on an array of 8 $\times$ 32GB NVIDIA V100 GPUs. The process utilized a learning rate of $1 \times 10^{-5}$ and a total batch size of 64 (32 for pre-training and 32 for ReFL).

\noindent\textbf{DPO.} Following the setting in~\cite{d3po}, we conducted a total of 400 epochs during the training process, utilizing a learning rate of $3 \times 10^{-5}$ and the Adam optimizer, alongside half-precision computation. This was conducted on a configuration comprising 8 $\times$ 32GB NVIDIA V100 GPUs.

\section{Cost of \our{} Construction}
\label{APP SEC: Cost}
One of the primary motivations for utilizing MLLMs as annotators is their ability to significantly reduce the cost of data construction compared to human annotators. Take construction process of the two largest existing human preference datasets as an example, during the construction process of Pick-a-Pic~\cite{kirstain2023pick}, approximately 6,394 web users participated in tagging images with their preferences. For the development of HPD v2~\cite{hpsv2}, a total of 57 high-quality annotation experts were employed and trained to construct preference labels. Moreover, these annotators were required to meticulously adhere to the annotation standards provided by the system throughout the labeling process. Thus, it is evident that obtaining large-scale humans preference annotation is time-consuming, resource-intensive, and laborious, which hinders the progress of related research. 

In contrast, employing MLLMs for annotation can effectively overcome these limitations. 
Utilizing the construction process of \our{} as an example, each invocation of the \gptv{} API is capable of tagging four images with preference labels pertaining to a specific aspect (e.g., prompt-following aspect), meaning a single API call can generate $C^{2}_{4}$ preference ranking results in that aspect. Throughout the construction of \our{}, each \gptv{} API can accommodate approximately 10,000 requests per day, thus generating around 60,000 preference ranking results in a given aspect per day. In the specific construction process, we employed two APIs for parallel annotation, with the total annotation process taking approximately 15 days.

This efficiency and cost-effectiveness are significantly superior to using human expert annotations. Moreover, despite the minimal cost and high efficiency, the reliability and quality of the preference labels provided by MLLMs are not compromised.

\section{Prompt Instruction Templates}
\label{APP SEC: Prompt Instruction Templates}

\subsection{Prompt Polish Instruction}
\label{app: polish prompt}

\begin{figure*}[!th]
\centering
\begin{tcolorbox}[
    colback=white, 
    colframe=black, 
    title=\textbf{Prompt Polish Instruction}, 
    fonttitle=\bfseries, 
    arc=1mm, 
]

I will give you a description about an image. Remove modifiers from text that have nothing to do with the main content of the image, for example resolution, sharpness, light, image quality, authors and online platform, and describe it succinctly in one sentence.

\#\# Original description (text): \{INSERT DESCRIPTION HERE\}
\\
\\
Note: Please provide your assessment results in the following format:
\\
\\
\#\#\# Output (text): [insert the sentence you generated here]
\end{tcolorbox}
\label{fig: polish prompt template}
\end{figure*}

\clearpage
\subsection{Preference Instruction}
\label{app: Preference prompt}

\begin{figure*}[!th]
\centering
\begin{tcolorbox}[
    colback=white, 
    colframe=black, 
    title=\textbf{Preference Instruction for Prompt-Following}, 
    fonttitle=\bfseries, 
    arc=1mm, 
]

\textbf{Prompt-Following:}

Your role is to evaluate the prompt-following quality score between given image and the corresponding text (``Input'').
The four images given are independent, and should be evaluated separately and step by step.
\\
\\
\textbf{Scoring}: Rating outputs 1 to 5:
\begin{enumerate}

\item \textbf{Irrelevant}: No alignment.

\item \textbf{Partial Focus}: Addresses one aspect poorly.

\item \textbf{Partial Compliance}:

    - (1) Meets goal or restrictions, neglecting other.
    
    - (2) Acknowledges both but slight deviations.

\item \textbf{Almost There}: Near alignment, minor deviations.

\item \textbf{Comprehensive Compliance}: Fully aligns, meets all requirements.
\end{enumerate}

\# Format:

\#\# Input:
\\
Text: \{INSERT PROMPT HERE\}
\\
Image: 

\#\#\# Image 1 [INSERT IMAGE 1 HERE]

\#\#\# Image 2 [INSERT IMAGE 2 HERE]

\#\#\# Image 3 [INSERT IMAGE 3 HERE]

\#\#\# Image 4 [INSERT IMAGE 4 HERE]
\\
\\
Note: Please provide your assessment results in the following format:
\\
\\
\#\# Output

\#\#\# Output for Image 1

Rating: [Rating for Image 1]

Rationale: [Rationale for the rating in short sentences]

\#\#\# Output for Image 2

Rating: [Rating for Image 2]

Rationale: [Rationale]

\#\#\# Output for Image 3

Rating: [Rating for Image 3]

Rationale: [Rationale]

\#\#\# Output for Image 4

Rating: [Rating for Image 4]

Rationale: [Rationale]

\end{tcolorbox}
\label{prompt: prompt-following}
\end{figure*}

\begin{figure*}[!th]
\centering
\begin{tcolorbox}[
    colback=white, 
    colframe=black, 
    title=\textbf{Preference Instruction for Aesthetic}, 
    fonttitle=\bfseries, 
    arc=1mm, 
]

\textbf{Aesthetic:}

Your role is to evaluate the aesthetic quality score of given images ("Images") generated by the corresponding text (``Input'').
The four images given are independent, and should be evaluated separately and step by step. Note that the rating has nothing to do with image input order.

\textbf{Scoring}: Rating outputs 1 to 5:
\begin{enumerate}

\item \textbf{Bad}: Extremely blurry, underexposed with significant noise, indiscernible subjects, and chaotic composition.

\item \textbf{Poor}: Noticeable blur, poor lighting, washed-out colors, and awkward composition with cut-off subjects.

\item \textbf{Fair}: In focus with adequate lighting, dull colors, decent composition but lacks creativity.

\item \textbf{Good}: Sharp, good exposure, vibrant colors, thoughtful composition with a clear focal point.

\item \textbf{Excellent}: Exceptional clarity, perfect exposure, rich colors, masterful composition with emotional impact.
\end{enumerate}

\# Format:

\#\# Input:
\\
Text: \{INSERT PROMPT HERE\}
\\
Image: 

\#\#\# Image 1 [INSERT IMAGE 1 HERE]

\#\#\# Image 2 [INSERT IMAGE 2 HERE]

\#\#\# Image 3 [INSERT IMAGE 3 HERE]

\#\#\# Image 4 [INSERT IMAGE 4 HERE]
\\
\\
Note: Please provide your assessment results in the following format:
\\
\\
\#\# Output

\#\#\# Output for Image 1

Rating: [Rating for Image 1]

Rationale: [Rationale for the rating in short sentences]

\#\#\# Output for Image 2

Rating: [Rating for Image 2]

Rationale: [Rationale]

\#\#\# Output for Image 3

Rating: [Rating for Image 3]

Rationale: [Rationale]

\#\#\# Output for Image 4

Rating: [Rating for Image 4]

Rationale: [Rationale]

\end{tcolorbox}
\label{prompt: Aesthetic}
\end{figure*}

\begin{figure*}[!th]
\centering
\begin{tcolorbox}[
    colback=white, 
    colframe=black, 
    title=\textbf{Preference Instruction for Fidelity}, 
    fonttitle=\bfseries, 
    arc=1mm, 
]

\textbf{Fidelity:}

Your role is to evaluate the fidelity score of given images ("Images") generated by the corresponding text (``Input'').
Higher fidelity score denote the image should be true to the shape and characteristics that the object should have and will not be generated haphazardly. Here are some counter-examples of low fidelity images:

\begin{enumerate}
\item “Spider-Man” should only have two legs, but generating an image with more legs is incorrect.
\item “Unicorn” should only have one horn, but generating an image with multiple horns is incorrect.
\item Generating more or less than five fingers for a human hand is incorrect, as each human hand has only five fingers.
\end{enumerate}

\textbf{Fidelity Rating}: Rating outputs 1 to 5:
\begin{enumerate}

\item \textbf{No fidelity}: There are lots of obvious shape and characteristics error in the image, and seriously affects the degree of beauty or coordination of the image.

\item \textbf{Insignificant fidelity}: There are lots of obvious shape and characteristics obvious error in the image, but the overall image is coordinated and acceptable.

\item \textbf{Partial fidelity}: There are some obvious mistakes (less than 2) but no harm, the overall image is coordinated.

\item \textbf{Severe fidelity}: There are a little inconspicuous obvious shape and characteristics mistakes, the full image is coordinated and beautiful.

\item \textbf{Completely fidelity}: No image errors, beautiful and coordinated images.
\end{enumerate}

\# Format:

\#\# Input:
\\
Text: \{INSERT PROMPT HERE\}
\\
Image: 

\#\#\# Image 1 [INSERT IMAGE 1 HERE]

\#\#\# Image 2 [INSERT IMAGE 2 HERE]

\#\#\# Image 3 [INSERT IMAGE 3 HERE]

\#\#\# Image 4 [INSERT IMAGE 4 HERE]
\\
\\
Note [IMPORTANT]: The four images given are independent, and should be evaluated the fidelity score of each image separately and step by step. Note that the rating has nothing to do with image input order. 
Please provide your assessment results in the following format:
\\
\\
\#\# Output

\#\#\# Output for Image 1

Rating: [Rating for Image 1]

Rationale: [Rationale for the rating in short sentences]

\#\#\# Output for Image 2

Rating: [Rating for Image 2]

Rationale: [Rationale]

\#\#\# Output for Image 3

Rating: [Rating for Image 3]

Rationale: [Rationale]

\#\#\# Output for Image 4

Rating: [Rating for Image 4]

Rationale: [Rationale]

\end{tcolorbox}
\label{prompt: Fidelity}
\end{figure*}

\begin{figure*}[!th]
\centering
\begin{tcolorbox}[
    colback=white, 
    colframe=black, 
    title=\textbf{Preference Instruction for Harmlessness}, 
    fonttitle=\bfseries, 
    arc=1mm, 
]

\textbf{Harmlessness:}

Your role is to evaluate the harmlessness score of given images ("Images") generated by the corresponding text (``Input'').
Higher harmlessness score denotes images do not include inappropriate content such as pornography, privacy violations, violence, discrimination, or generally not safe for work (NSFW) themes.
The four images given are independent, and should be evaluated separately and step by step. Note that the rating has nothing to do with image input order.

\textbf{Harmlessness Rating}: Rating outputs 1 to 5:
\begin{enumerate}

\item \textbf{Unsafe}: Not suitable for general audiences, includes explicit sexual content, real violence, terror imagery, or privacy violations.

\item \textbf{Potentially Sensitive}: NSFW, with explicit violence, suggestive themes, graphic medical content, but no pornography or severe violence.

\item \textbf{Moderately Safe}: Suitable for adults, may not suit children. Includes moderate violence, implied nudity, and political or social commentary. No explicit sexual content or severe violence.

\item \textbf{Generally Safe}: Safe for most, with mild thematic elements. Includes non-sexual nudity in art or history, minimal violence, and mild satire.

\item \textbf{Highly Safe}: Images are free from any offensive, sensitive, or potentially harmful content, with no traces of violence, nudity, or any form of hate speech. They are suitable for all.
\end{enumerate}

\# Format:

\#\# Input:
\\
Text: \{INSERT PROMPT HERE\}
\\
Image: 

\#\#\# Image 1 [INSERT IMAGE 1 HERE]

\#\#\# Image 2 [INSERT IMAGE 2 HERE]

\#\#\# Image 3 [INSERT IMAGE 3 HERE]

\#\#\# Image 4 [INSERT IMAGE 4 HERE]
\\
\\
Note [IMPORTANT]: Please provide your assessment results in the following format:
\\
\\
\#\# Output

\#\#\# Output for Image 1

Rating: [Rating for Image 1]

Rationale: [Rationale for the rating in short sentences]

\#\#\# Output for Image 2

Rating: [Rating for Image 2]

Rationale: [Rationale]

\#\#\# Output for Image 3

Rating: [Rating for Image 3]

Rationale: [Rationale]

\#\#\# Output for Image 4

Rating: [Rating for Image 4]

Rationale: [Rationale]

\end{tcolorbox}
\label{prompt: Harmlessness}
\end{figure*}

\end{document}